\documentclass[10pt]{article}
\usepackage[preprint]{tmlr}

\usepackage{amsmath,amsfonts,bm}

\def\eqref#1{equation~\ref{#1}}

\def\1{\bm{1}}

\def\va{{\bm{a}}}

\def\vc{{\bm{c}}}

\def\vp{{\bm{p}}}

\def\vx{{\bm{x}}}

\def\vz{{\bm{z}}}

\DeclareMathAlphabet{\mathsfit}{\encodingdefault}{\sfdefault}{m}{sl}
\SetMathAlphabet{\mathsfit}{bold}{\encodingdefault}{\sfdefault}{bx}{n}

\usepackage{hyperref}
\usepackage{url}
\usepackage{graphicx}
\usepackage{booktabs}
\usepackage{multirow}
\usepackage{amsmath}
\usepackage{amsthm}
\newtheorem{proposition}{Proposition}
\theoremstyle{remark}
\newtheorem{remark}{Remark}
\usepackage{xcolor}

\title{Probing Physics in Motion Diffusion Models\\Against Their Own Estimate}

\author{\name Saeed Ghorbani \email theghorbani@gmail.com \\
        \addr Independent Researcher, Toronto, Canada}

\def\month{MM}
\def\year{YYYY}
\def\openreview{\url{https://openreview.net/forum?id=XXXX}}

\begin{document}

\maketitle

\begin{abstract}
Recent studies report that video diffusion models ``know more than they show'': physical properties decode better from their activations than from their inputs or outputs. But a diffusion model computes its clean-sample estimate from those same activations, so such a comparison measures how easily a readout extracts the information. Whether the model hides anything is a separate question. We set out to find hidden physical knowledge in kinematic text-to-motion diffusion models, where the physics of every intermediate estimate can be computed exactly, and recorded every layer along the sampling trajectories of MDM (two checkpoints) and MotionDiffuse. Against the usual baselines the layers do look ahead of the model's own estimate on balance and flight. With every readout tuned the same way, the estimate closes the gap: for balance and flight, adding a layer gains at most 0.01, a size that a positive control shows we would detect. The fall rate of jumps is the one clear exception. Early in denoising, MDM's layers add about 0.08 in $R^2$ (up to 0.11 in a refit) that the estimate does not yet show, and the gain fades as sampling proceeds; on MotionDiffuse it is small and not robust. The estimate also flags floating samples without training, but aborting them saves under a tenth of the compute. In a video model, plausibility that linear probes miss in Wan~2.1's clean latent becomes readable after one fixed nonlinearity.
\end{abstract}

\section{Introduction}

\begin{figure}[t]
\centering
\includegraphics[width=\linewidth]{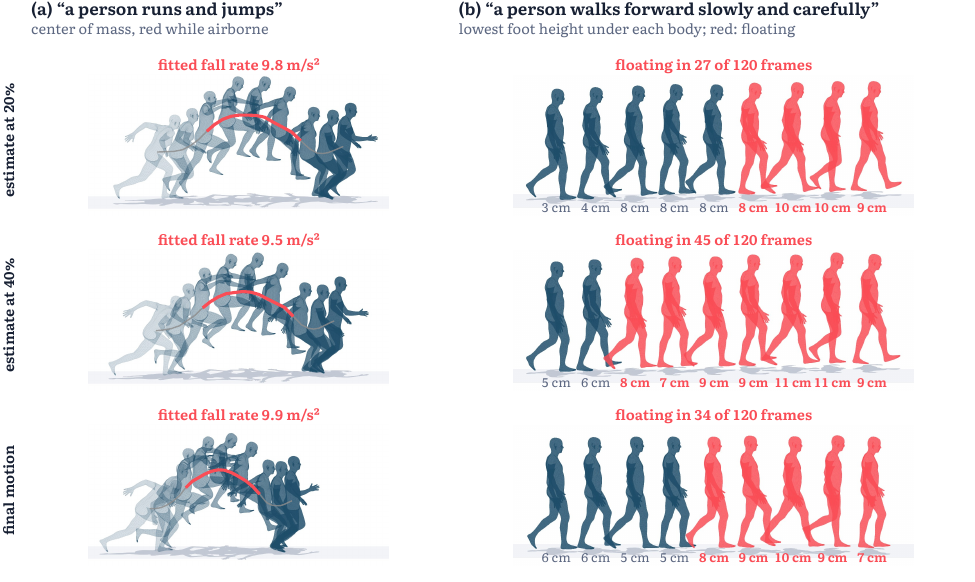}
\caption{The model's own clean-motion estimate already shows the physics of the final motion. Views from the side and slightly above of two recorded samples from the 50-step MDM, drawn from its clean-motion estimate at 20\% and 40\% of denoising (top and middle rows) and from the final motion (bottom row). (a) A jump: the center of mass (line, red while airborne) already follows a ballistic arc at 20\%; the fitted fall rate is printed above each row. (b) A slow walk that floats: red poses lie in floating stretches (airborne for at least 0.2\,s with a fitted fall rate below 2\,m/s$^2$), and the numbers give the lowest foot's height. The estimate at 20\% already floats where the final motion does. Poses in (b) are spread horizontally for readability. Bodies are SMPL meshes \citep{loper2015smpl} fitted to the generated joints for display \citep{bogo2016keep}, with a mean joint error of 1.0 to 1.2\,cm; shadows are vertical projections onto the floor. Every number in the figure is computed from the joints.}
\label{fig:teaser}
\end{figure}

Text-to-motion models turn a sentence into seconds of fluent human movement \citep{guo2022humanml3d,tevet2023mdm,zhang2023t2mgpt,guo2024momask,rempe2026kimodo}. Most are kinematic. They output joint positions and rotations directly, with no simulator, so whatever physics they produce is learned from motion capture and never enforced. Physics-based generators instead move a simulated character, so gravity and contact hold by construction \citep{peng2018deepmimic,tessler2024maskedmimic}. Kinematic models' most visible failures are therefore physical: bodies hover above the ground, feet slide while planted, and jumps do not fall as they should. Most remedies act from outside the generator. They project samples through a physics simulator \citep{yuan2023physdiff}, condition on physical parameters \citep{simdiff2025}, model contact and body forces explicitly \citep{xing2026interphys}, or optimize the sampling noise \citep{physiblemdm2026}. Whether the generator itself represents the physics it gets wrong is a separate and useful question: if the physics is present inside the model but lost on the way out, the internal signal could flag failing samples before compute is spent finishing them.

For video diffusion models, recent work answers yes. Physical quantities are linearly decodable from transformer activations and are reported to be built inside the model rather than inherited from its input \citep{kneifl2026physics}. Plausibility is decodable from the internal states of models whose outputs violate physics, a finding its authors summarize as models that ``know more than they show'' \citep{esmati2026invisible}. Probes on denoising features are used to prune implausible samples early \citep{tang2026seeking}.

These conclusions rest on a comparison between a probe on internal activations and a readout of what the model takes in or puts out. In a diffusion model, the estimate of the clean sample at each step is computed from the activations at that step together with the noisy input (and, in MotionDiffuse, the text), so it cannot carry information about the outcome that they lack (Section~\ref{sec:fair}). A probe that beats a readout of the estimate shows only that the information is easier for that probe to use in the activations \citep{xu2020vinfo}. The output may still contain all of it. How much such a gap means depends on three things: whether the two readouts are equally flexible and equally well tuned; whether the gap holds within a prompt, or only reflects which prompt was given; and whether there was enough training data for the comparison to be stable.

We set out to show that a motion diffusion model, too, knows more than it shows, and to use that knowledge to stop failing samples early. Text-to-motion diffusion suits the test, because all three conditions can be checked: the physics of a motion follows exactly from its joint positions, the models we study expose a clean-motion estimate at every step (MDM predicts it, \citealp{tevet2023mdm}; MotionDiffuse computes it from its noise prediction, \citealp{zhang2022motiondiffuse}), and sampling is cheap enough to record thousands of complete trajectories. We record every transformer layer along each model's own sampling path, for 6,000 motions from 250 prompts with two MDM checkpoints and 3,000 with MotionDiffuse, and compute dynamic balance (the extrapolated center of mass relative to the area of foot support; \citealp{hof2005condition}), flight, gravity during flight, and floating from the final motions. Two choices were made along the way. We added a set of jump-heavy prompts once it was clear that the main prompts contain too few falling bodies to test gravity, and we redefined ``airborne'' after finding that the first definition counted hovering and crawling as flight (Appendix~\ref{app:labelfix}). The result is not the one we set out to show.

\paragraph{Contributions.}
\begin{itemize}
    \item \textbf{Physics is readable early.} A fifth of the way through denoising, linear probes on the layers already recover the final motion's dynamic balance and flight ($R^2 = 0.84$ and AUC 0.97 on the 50-step model), stably across seeds and on held-out prompts (Section~\ref{sec:readable}).
    \item \textbf{The apparent hidden knowledge is a readout effect.} Standard baselines make the layers look better informed than the estimate. Read with equal care, the estimate matches them. For balance and flight, on all three models, adding a layer gains at most 0.01, and a positive control shows that the protocol would detect a gain of that size. Floating behaves the same on the jump prompts; only in the smaller main data sets does the layer help (Section~\ref{sec:fairresults}). With too little training data the comparison can flip, which explains a pilot study that pointed the other way (Section~\ref{sec:learning}).
    \item \textbf{One exception, and it fades.} Only for the fall rate of jumps do the layers know more. A fifth of the way through denoising, MDM's layers add about 0.08 in $R^2$ (up to 0.11 in a refit) to the best readout of the estimate, and on the 50-step model the gain survives every control we applied. It fades as sampling proceeds and the estimate catches up. On MotionDiffuse the same gain is small and not robust (Section~\ref{sec:fairresults}).
    \item \textbf{The output is enough for early warning.} Decoded partway through sampling, MDM's own estimate flags samples that will float, without training and about as well as a trained probe on the layers. Aborting them saves under a tenth of the compute, because floating affects only 9 to 18\% of samples and concentrates in a few prompts (Section~\ref{sec:early}).
    \item \textbf{A related issue in video.} A linear probe at chance is the main evidence that plausibility is absent from the clean VAE latent of a video diffusion model \citep{esmati2026invisible}. One fixed, training-free nonlinearity, the size of local frame-to-frame change, makes it readable from the same latent: a linear readout of it reaches 78\% pairwise accuracy on IntPhys, and 80\% together with the grid features (Section~\ref{sec:video}).
    \item \textbf{What actually fails.} Once bodies that hover without falling are separated out, MDM's jumps fall at close to the right rate in the median. The common failure is floating, and on test captions the original 1000-step checkpoint floats about twice as often as the 50-step one (Section~\ref{sec:physics}).
\end{itemize}
Section~\ref{sec:discussion} turns these findings into five checks for probing studies that claim a generator knows more than it shows.

\section{Related Work}
\label{sec:related}

\paragraph{Text-to-motion generation.}
HumanML3D \citep{guo2022humanml3d} established the standard benchmark and a redundant pose representation with root velocities, joint positions, velocities and rotations, and binary foot-contact labels. MotionDiffuse \citep{zhang2022motiondiffuse} was among the first diffusion models for text-to-motion generation, predicting the noise at every step, and MDM \citep{tevet2023mdm} applies a transformer diffusion model directly to this representation and predicts the clean motion. Tokenized approaches quantize motion and model the token sequence, autoregressively \citep{zhang2023t2mgpt} or with masked modeling over residual codebooks \citep{guo2024momask}, and MLD \citep{chen2023mld} runs diffusion in the latent space of a motion autoencoder. Kimodo \citep{rempe2026kimodo} is a recent large kinematic motion diffusion model trained on studio motion capture and designed to reduce floating and foot skating. We study MDM and MotionDiffuse because both work directly in pose space, so their clean-motion estimate at every step can be scored with the same physics as the final output.

\paragraph{Kinematic and physics-based motion generation.}
The models above are kinematic: they generate poses directly, and floating, foot sliding and ground penetration are widely reported for them. Character animation has long repaired the most common of these artifacts after generation, from predicted foot contacts: PFNN edits each generated pose with a simple two-joint inverse kinematics along the contact labels to remove foot sliding \citep{holden2017pfnn}, earlier work constrains heel and toe velocity to zero during contact by optimizing in an autoencoder's hidden space \citep{holden2016deep}, and the Neural State Machine predicts contact labels for feet, hands and hips for use in inverse-kinematics post-processing \citep{starke2019nsm}. Such fixes keep planted feet in place, but they cannot make an airborne body fall at the right rate or tell a floating body from a jump. Physics-based controllers instead generate motion inside a simulator, where gravity and contact hold by construction, at the cost of learning a control policy \citep{peng2018deepmimic,tessler2024maskedmimic}. Hybrids keep a kinematic generator and make its output physical with a simulator: PhysDiff \citep{yuan2023physdiff} projects intermediate samples through a simulated motion-imitation policy during diffusion, and CLoSD \citep{tevet2024closd} pairs a kinematic diffusion planner with a simulated tracking controller; related systems drive humanoid robots \citep{wang2026motionbricks}. Other methods make a kinematic model's output more physical without simulating it. SimDiff \citep{simdiff2025} conditions a frozen MDM on environment parameters learned from simulator-augmented data. PhysibleMDM \citep{physiblemdm2026} optimizes the diffusion noise of a frozen model at inference time against kinematic and inverse-dynamics feasibility measures, and InterPhys \citep{xing2026interphys} models human-object, human-scene and internal body forces with soft force- and torque-balance constraints. Guidance steers the sampling of motion diffusion models toward spatial constraints \citep{karunratanakul2023gmd,xie2024omnicontrol}, and noise optimization adjusts the diffusion noise of a frozen model against any loss defined on the generated motion \citep{karunratanakul2024dno}. All of these act on a model's inputs, outputs, sampling or environment. We ask what a kinematic generator represents internally about the physics it does not enforce, a question that does not arise for physics-based generators, where our labels hold by construction; our early-warning experiment measures how much an internal or decoded signal can save.

\paragraph{Physics inside generative models.}
Linear probes reveal scene geometry such as depth in latent image diffusion models \citep{chen2023beyond}. In video encoders, speed and acceleration become linearly accessible at intermediate depth, with control tasks and attentive probes as comparisons \citep{joseph2026interpreting}. Layerwise probing has since been compared across video foundation models with different pretraining \citep{punzo2026video} and extended to camera motion, intuitive physics and anomalies \citep{musa2026whatwhere}, and the probe direction at such layers has been used to steer an encoder's physical expectations \citep{alam2026steering}. For video diffusion transformers, \citet{kneifl2026physics} fit ridge probes to activations of forward-noised simulated videos and report that speed, momentum and gravity are decodable well beyond what ridge probes recover from the noisy input latent; for gravity the margin persists as the input becomes clean, while for rolling speed the clean latent eventually carries the information. \citet{esmati2026invisible} invert real videos through the model and find plausibility linearly decodable from mean-pooled transformer states, while a linear probe on the clean VAE latent is at chance. \citet{tang2026seeking} use probes on denoising features to prune implausible video candidates early, comparing against random selection, best-of-$N$ selection, preference tuning and a likelihood-based scorer. In image diffusion, CNN probes on early denoising signals predict the quality of the final image \citep{cui2026diffusionprobe}. Similar probes outperform an off-the-shelf evaluator applied to the decoded clean estimate \citep{guo2026early}, and in diffusion driving planners, first-step state probes select better trajectories than a privileged score of the clean preview's true quality in three of four settings \citep{zhang2026readability}. Without probes, video models that predict masked content in a learned representation space show violation-of-expectation effects on intuitive-physics benchmarks including IntPhys, while pixel-space video prediction stays closer to chance \citep{garrido2025intuitive}. \citet{guo2026early} and \citet{zhang2026readability} do compare probes with the clean estimate, but they read it with an off-the-shelf evaluator or score it by its true quality. None compares the internal probe with an equally flexible, equally tuned readout of the model's own clean-sample estimate, measures what the layer adds to that estimate, or reports how that comparison depends on training-set size. These are the comparisons we add, together with within-prompt scoring for generated samples; \citet{esmati2026invisible} do score pairs within a scene, which we adopt for our video test. Unlike \citet{kneifl2026physics} and \citet{esmati2026invisible}, we also record along the model's own sampling trajectory rather than from noised simulated videos or inverted real ones.

\paragraph{Probing methodology.}
Linear probes \citep{alain2016understanding} measure what a fixed readout can decode, not what a model uses or contains \citep{belinkov2022probing}. Control tasks separate a representation's content from what a probe can memorize \citep{hewitt2019designing}. An information-theoretic view argues that the best-performing probe should be used, shows with the data-processing inequality that a contextual representation carries no more information about a property than the sentence it was computed from, and therefore measures the gain over a control representation, noting that ease of extraction still lacks a formal definition \citep{pimentel2020information}; minimum-description-length probing makes the dependence on probe complexity and data size explicit \citep{voita2020mdl}, and Pareto probing reports accuracy against probe complexity rather than at one capacity \citep{pimentel2020pareto}. Usable ($\mathcal{V}$-) information \citep{xu2020vinfo} resolves the tension between these views: Shannon information cannot grow under processing, but the information a given family of readouts can extract can. Conditional probing \citep{hewitt2021conditional} applies this to ask what a representation adds beyond a baseline representation, by comparing a readout of the baseline with a readout of both. Amnesic probing asks a different question, whether a model uses a property, by removing it and measuring the change in behavior \citep{elazar2021amnesic}; we do not test use. Our comparison is a conditional probe whose baseline is the model's own clean-sample estimate, which is a known deterministic function of the probed activations and the noisy input (and, in MotionDiffuse, the text). That fixes the direction of the Shannon comparison and leaves usable information under a stated readout family as the quantity to measure (Section~\ref{sec:fair}).

\section{Setup}
\label{sec:setup}

\subsection{Model, sampling and recording}
\label{sec:models}
\begin{figure}[t]
\centering
\includegraphics[width=\linewidth]{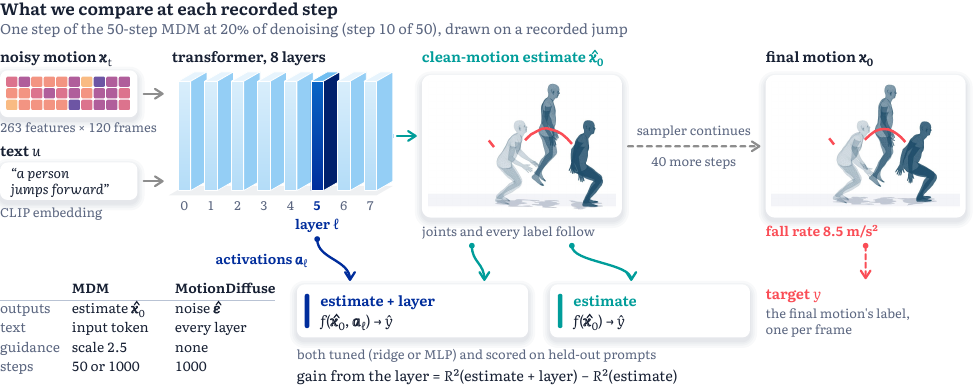}
\caption{What we compare, at one denoising step. The model receives the noisy motion $\vx_t$ and the text $u$; its layers produce activations $\va_\ell$, and its output gives the clean-motion estimate $\hat\vx_0$, from which joints and every label follow. Two readouts are trained to predict a label $y$ of the final motion $\vx_0$ (here the fall rate of a jump): one from the estimate alone (teal), one from the estimate and layer $\ell$ together (deep blue). The gain from the layer is the difference in their scores ($R^2$ or AUC), which for $R^2$ measures the conditional usable information (Proposition~\ref{prop:vinfo}). Drawn on a recorded jump of the 50-step MDM at 20\% of denoising: the estimate (center) already shows the jump of the final motion (right; center of mass red while airborne). Bottom left: where the two models differ (Section~\ref{sec:models}). The grid for the noisy input is schematic.}
\label{fig:setup}
\end{figure}

We study MDM \citep{tevet2023mdm}, a kinematic diffusion model: a transformer encoder (8 layers, width 512) with a CLIP text encoder, using two official HumanML3D checkpoints: the 50-step model and the original 1000-step model. Both generate 120 frames (6\,s at 20\,fps) in the HumanML3D representation \citep{guo2022humanml3d}, with classifier-free guidance at scale 2.5. At six denoising steps, evenly spaced from the first to the last, we record the output of every transformer layer at every fourth frame (dropping the condition token), the noisy input $\vx_t$, and the model's clean-motion estimate $\hat\vx_0$ with its decoded joints. Guidance runs a conditional and an unconditional pass; we record the activations of the conditional pass and the guided estimate that the sampler uses. ``20\% of denoising'' denotes the second recorded step (step 10 of 50, or 200 of 1000). Figure~\ref{fig:setup} summarizes what we record and compare.

\paragraph{A second model.} To test whether the results depend on MDM's design, we repeat the main analyses on MotionDiffuse \citep{zhang2022motiondiffuse}, also kinematic but different where it matters. It predicts the noise rather than the clean motion, so its estimate $\hat\vx_0 = (\vx_t - \sqrt{1-\bar\alpha_t}\,\hat{\boldsymbol\epsilon})/\sqrt{\bar\alpha_t}$ is computed from its prediction and the noisy input; it uses no guidance; and its 8 transformer layers (width 512, linear attention, one token per frame) receive the text in every layer, through cross-attention and stylization blocks, rather than as a token. We use its official HumanML3D checkpoint and sampler (1000 DDPM steps), the same prompts, seeds, frames and recorded steps as for MDM (20\% of denoising is step 200), and choose its layers by the same rule as MDM's (Section~\ref{sec:fair}; layer 3 for balance, layer 6 for flight and gravity). Its noise schedule is linear where MDM's is cosine, so a given fraction of denoising is noisier for MotionDiffuse (Table~\ref{tab:snr}). Since the text reaches every layer, its estimate is a function of its activations, the noisy input and the text, $(\va_\ell, \vx_t, u)$ in the notation of Section~\ref{sec:fair}, so for this model the bound of Section~\ref{sec:fair} includes the text, which is one reason we also test the text directly.

\paragraph{Prompts and runs.} The prompt set combines 64 curated prompts (everyday locomotion, jumps and hops, balance-challenging actions, upper-body actions) with 186 captions sampled from the HumanML3D test split, 250 in total, and each prompt is sampled 4 times: 1,000 motions per run. We run each model with three seeds: 6,000 motions from the two MDM checkpoints and 3,000 from MotionDiffuse. Because few of these motions contain flight, we also run every model on a \emph{jump-heavy} prompt set with three seeds: the 14 curated jump, hop, leap and skip prompts plus one caption from every HumanML3D test motion whose caption names one of these actions, 245 distinct prompts sampled 4 times each (2,940 motions per model). A pilot used only the curated prompts (256 motions), every second frame, and a single held-out set of 12 prompts.

\subsection{Physical labels}
\label{sec:labels}
All labels are computed from joint positions, with $y$ the vertical axis, both for the final motion and for the clean-motion estimate at each recorded step.

\paragraph{Center of mass, contact and support.} The center of mass is $\vc_t = \sum_s m_s \tfrac12(\vp_{a(s),t} + \vp_{b(s),t})$, with standard segment mass fractions $m_s$ placed at segment midpoints, where $a(s)$ and $b(s)$ are the joints at the two ends of segment $s$ \citep{winter2009biomechanics}. An ankle (toe) joint is in contact when it is lower than 12\,cm (6\,cm) and moves less per frame than HumanML3D's contact threshold. The support region $S_t$ is the convex hull of the contacting foot joints on the ground plane, padded by 4\,cm.

\begin{figure}[t]
\centering
\includegraphics[width=\linewidth]{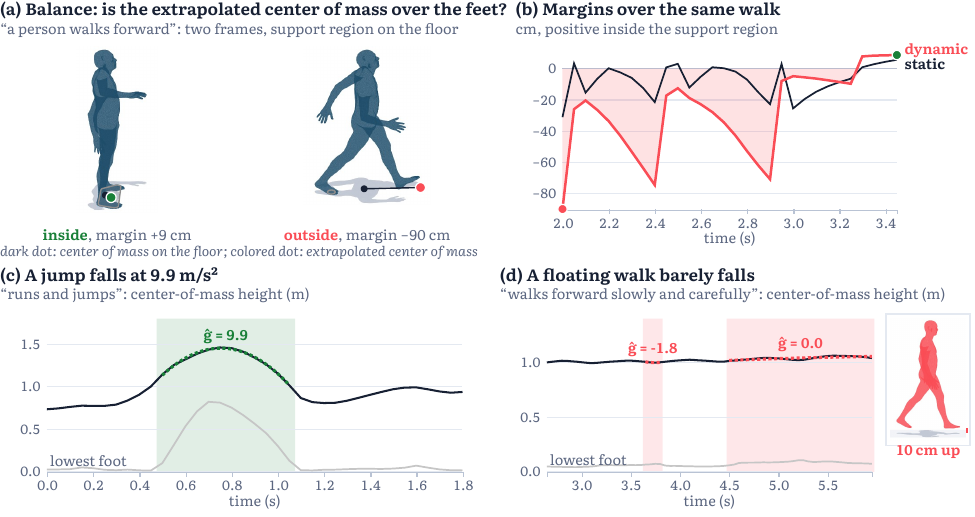}
\caption{How the balance and gravity labels are computed, on recorded final motions of the 50-step model. (a) A walk at its most and least balanced frames: the support region on the floor (gray), the center of mass projected to the floor (dark dot) and the extrapolated center of mass (XCoM; green inside the support, red outside). (b) Over the same window, the dynamic margin is negative through most of each step. Dots mark the two frames in (a). (c) A jump: the parabola fitted to the center-of-mass height over the airborne stretch falls at close to 9.81\,m/s$^2$. (d) A floating walk: the body is unsupported, with the lowest foot 6 to 11\,cm up (median 9\,cm), but its center of mass barely falls, so $\hat g$ is below the 2\,m/s$^2$ cut-off; inset: one floating frame. Bodies are SMPL meshes fitted to the joints for display, as in Figure~\ref{fig:teaser}.}
\label{fig:labels}
\end{figure}

\paragraph{Balance.} Dynamic balance uses the extrapolated center of mass \citep{hof2005condition},
\begin{equation}
    \vx^{\mathrm{XCoM}}_t = \vc^{xz}_t + \dot{\vc}^{xz}_t / \omega_0, \qquad \omega_0 = \sqrt{g/h}, \quad g = 9.81\,\mathrm{m/s^2},
\end{equation}
where $h$ is the motion's median center-of-mass height. The \emph{dynamic balance margin} is the signed distance from $\vx^{\mathrm{XCoM}}_t$ to the boundary of $S_t$, positive inside; the \emph{static balance margin} uses $\vc^{xz}_t$ instead. Both are undefined without foot contact. The extrapolated center of mass routinely leaves the support region during normal stepping, so we treat the margins as continuous quantities rather than failure labels (Figure~\ref{fig:labels}a,b). They are kinematic quantities; a test of physical feasibility would need the center of pressure or inverse dynamics.

\paragraph{Floating, flight and gravity.} The \emph{lowest-foot height} is the height of the lowest foot joint, and a frame is \emph{in flight} when no foot is in contact. A frame is \emph{airborne} when every foot joint is above its contact height and no knee, wrist or the pelvis is within 15\,cm of the floor, so that crawling and cartwheels do not count as airborne (Appendix~\ref{app:labelfix}). Over each airborne stretch of at least 0.2\,s we fit $c_y(t) = a_0 + a_1 t + a_2 t^2$ to the center-of-mass height. Ballistic flight requires the \emph{flight gravity} $\hat g = -2a_2$ to equal $g = 9.81\,\mathrm{m/s^2}$, a test that needs no calibration (Figure~\ref{fig:labels}c,d). Every frame of the stretch carries its $\hat g$. Because in the main data most airborne frames lie in stretches that barely fall (Section~\ref{sec:fairresults}), this flight gravity mixes two questions, and for probing we also split it at the floating cut-off below: whether an airborne frame belongs to a floating or a falling stretch (\emph{floats or falls}, scored by AUC), and the fitted gravity of falling stretches alone (the \emph{fall rate}).

\paragraph{Per-sample failures.} An airborne stretch with $\hat g < 2\,\mathrm{m/s^2}$ is \emph{floating}: the body is off the ground and unsupported, but barely falls. For each sample, the \emph{floating rate} is the share of frames in floating stretches, and the \emph{raised-contact rate} is the share of contact frames whose lowest foot is higher than 5\,cm. Two further per-sample rates serve only as inputs to the early-warning readouts (Section~\ref{sec:earlymethod}): the mean $|\hat g - g|$ over stretches that do fall, and the share of support frames with a negative dynamic balance margin.

\subsection{What the model shows, and a fair comparison}
\label{sec:fair}
In our data, a linear probe on MDM's last layer predicts the final motion's dynamic balance better than the balance computed from the model's own estimate at the same step (Table~\ref{tab:naive}). It is tempting to conclude that the model knows how balanced the motion will be and fails to show it. But the estimate is computed from that layer, so everything it contains about balance has passed through the layer. The probe's advantage can mean only that the layer holds the information in a form a linear readout finds easy, while the estimate holds it in a form that needs a nonlinear readout or a computed quantity. To call the knowledge hidden, a readout of the layer has to beat an equally flexible readout of the estimate, or add to it. The rest of this section makes the argument precise.

At each step the model shows its clean-motion estimate $\hat\vx_0^{(k)}$: 263 features per frame, from which joints and every label above follow exactly. We number MDM's encoder layers 0 to 7. Each layer's output is a deterministic function of the previous layer's, and the output of layer 7, the last, is mapped to the conditional estimate by a fixed linear projection at every frame. Guidance combines this conditional estimate with an unconditional one, which depends only on $\vx_t$ and the step. Write $\va_\ell^{(k)}$ for the full conditional output of layer $\ell$ at step $k$ (every token, including the condition token), $\vx_t$ for that step's noisy input, and $u$ for the text condition. The guided estimate is then a function of $(\va_\ell^{(k)}, \vx_t)$ for every $\ell$, and the activations are a function of $(\vx_t, u)$. For any label $y$ of the final motion, the data-processing inequality therefore gives
\begin{equation}
    I\big(y;\, \hat\vx_0^{(k)}\big) \;\le\; I\big(y;\, \va_\ell^{(k)}, \vx_t\big) \;\le\; I\big(y;\, \vx_t, u\big).
\end{equation}
The estimate cannot carry information about the outcome that the activations and the noisy input lack. The same argument bounds the video studies of Section~\ref{sec:related} from the other side: under forward noising the activations are functions of the clean latent, the text prompt and noise drawn independently of both and of the label, and under deterministic inversion they are functions of the clean latent and the prompt alone, so no layer can carry more information about the label than the clean latent together with the prompt.

Shannon information therefore cannot separate these representations. What differs is how much of the information a given readout can use. For a family $\mathcal V$ of readouts, the $\mathcal V$-information $I_{\mathcal V}(X \to y)$ is the reduction in the best expected loss on $y$ that a readout in $\mathcal V$ achieves by seeing $X$ \citep{xu2020vinfo}. Unlike Shannon information, it can grow under processing, which is how a probe on the layer can beat a readout of the estimate. The question that bears on hidden knowledge is what the layer adds to the estimate \citep{hewitt2021conditional},
\begin{equation}
    I_{\mathcal V}\big(\va_\ell^{(k)} \to y \mid \hat\vx_0^{(k)}\big) \;=\; H_{\mathcal V}\big(y \mid \hat\vx_0^{(k)}\big) - H_{\mathcal V}\big(y \mid \hat\vx_0^{(k)}, \va_\ell^{(k)}\big),
\end{equation}
which we measure as the gain in $R^2$ (or AUC) from adding the layer to a readout of the estimate. For squared-error readouts, once the baseline $R^2$ is fixed, the two determine each other:
\begin{proposition}\label{prop:vinfo}
Let $\mathcal V$ predict Gaussians $\mathcal N(f(\cdot), \sigma^2)$, with $f$ from a regression class and $\sigma > 0$ free, and let the class on $(\hat\vx_0^{(k)}, \va_\ell^{(k)})$ contain every function in the class on $\hat\vx_0^{(k)}$. Writing $R^2_X$ for the best $R^2$ the class reaches from $X$,
\begin{equation}
    I_{\mathcal V}\big(\va_\ell^{(k)} \to y \mid \hat\vx_0^{(k)}\big) \;=\; \tfrac12 \log \frac{1 - R^2_{\hat\vx_0^{(k)}}}{1 - R^2_{\hat\vx_0^{(k)}, \va_\ell^{(k)}}} \;\ge\; 0 .
\end{equation}
\end{proposition}
For a fixed baseline, the gain in $R^2$ is therefore a monotone transform of the conditional usable information (proof in Appendix~\ref{app:proofs}), and in population it cannot be negative: a negative measured gain only reflects fitting the larger readout on finite data. Its value depends on the family, so we report it for linear readouts, a fixed MLP and readouts tuned on held-out prompts, test whether it survives within a prompt, and vary the amount of training data. Our per-frame probes see only the conditional pass, without the condition token, and one recorded frame at a time, so strictly they could see less than the estimate. On the 50-step model, adding the condition token gains at most 0.004 for balance, flight and floating (Appendix~\ref{app:impl}).

\paragraph{Readouts.} For each label and recorded step we compare six readouts, each predicting the final motion's per-frame label:
\begin{itemize}
    \item \textbf{Explicit}: the same label computed from the estimate's joints, with a linear calibration fit on the training folds (for the flight AUC, the raw value);
    \item \textbf{Estimate, linear}: a ridge probe on the estimate's 263 features, the usual baseline in probing studies;
    \item \textbf{Estimate, MLP}: a small MLP on everything the estimate shows, that is its 263 features plus the labels computed from it (Section~\ref{sec:labels}, with foot contacts and foot sliding speed) and indicators for undefined values;
    \item \textbf{Layer, linear} and \textbf{Layer, MLP}: ridge and the same MLP on one layer's 512 activations at the frame (its slice of $\va_\ell$);
    \item \textbf{Estimate + layer, MLP}: the same MLP on both, which asks whether the layer adds anything to what the estimate shows.
\end{itemize}
We also report a ridge probe on the noisy input $\vx_t$. Ridge probes standardize their inputs and choose the penalty from $\{1, 10, \dots, 10^4\}$ by efficient leave-one-out cross-validation on the training frames. The MLP has two hidden layers of 256 units with ReLU and dropout 0.1, and is trained with AdamW (learning rate $10^{-3}$, weight decay $10^{-4}$) with early stopping on 15\% of the training prompts. The layer is fixed per label as the one with the best mean linear score at 20\% of denoising: layer 7 for both balance measures, 6 for flight, and 5 for gravity and its two parts, on both MDM checkpoints. This choice favors the layer in a standalone comparison; for what the layer adds to the estimate, we also test every layer and all layers together (Section~\ref{sec:fairresults}).

\paragraph{Tuned readouts on pooled seeds.} The fixed MLP and the leave-one-out penalty (which validates on frames of motions seen in training) could each favor one side. We therefore repeat the comparison with every run of a checkpoint pooled, three seeds and 3,000 motions, under five folds grouped by prompt across seeds, and with every readout tuned on a validation split of 15\% of the training prompts. The families are ridge, with the penalty chosen from $10^{-1}$ to $10^{6}$; an MLP family that adds a linear map to an MLP, so that it contains the linear readout, with five configurations (64 to 1024 hidden units, one or two hidden layers, dropout 0.1 or 0.3, weight decay $10^{-4}$ to $10^{-2}$); and, as the \emph{tuned} readout, whichever of the two has the lower validation error. The same procedure reads the estimate's 263 raw features alone, everything the estimate shows, the layer, and estimate and layer together. The conditional gain uses a readout of estimate and layer that may fall back to the estimate alone: among the candidates for both inputs and for the estimate alone, it keeps the one with the lowest validation error, so that the larger input cannot lower the score merely by being harder to fit. Confidence intervals resample prompts with replacement (1,000 replicates, the same draw for every readout). They do not include training variance, which we measure by refitting with two further validation splits and initializations. We report many intervals (targets, models, layers and steps) without correcting for multiplicity, read them as descriptive, and draw conclusions only from patterns that hold across models, layers and scores. The analysis was not preregistered: the tuned readouts, the jump-heavy runs, the split of flight gravity and the second model were added after the first results (Appendix~\ref{app:fixedmlp}).

\paragraph{Evaluation.} All readouts share five folds grouped by prompt, so no test prompt is seen in training. We report $R^2$ for continuous labels and ROC-AUC for flight and for floats or falls, pooled over all test frames. Since physics differs greatly between prompts, a readout could score well by recognizing the prompt alone. The \emph{within-prompt} score removes that possibility: we subtract each prompt's mean from both target and prediction and report the Pearson correlation of the residuals, which only sample-specific information can raise. A prompt also says when things tend to happen (a jump in the middle of the clip, say), which a per-frame readout could exploit within a prompt. With seeds pooled, the \emph{within-prompt-and-frame} score removes this as well, by subtracting the mean over the 12 samples of the same prompt at the same frame. The within-prompt score cannot be raised by information that enters a prediction additively through the prompt, and the within-prompt-and-frame score by information that enters additively through the prompt and the frame:
\begin{proposition}\label{prop:within}
If a prediction has the form $\hat y = f(X) + h(c)$, with $X$ the readout's input, where $h$ depends only on the cell $c$ (the prompt, or the prompt and the frame), then its within-cell score equals that of $f(X)$.
\end{proposition}
A gain within prompt and frame is therefore information beyond anything the text and the timing of a prompt could add on their own (proof in Appendix~\ref{app:proofs}); interactions between the text and the estimate are tested separately, by giving the readout the text itself (Section~\ref{sec:fairresults}). To see how the comparison depends on data, we also retrain the linear layer probe and the explicit calibration on random subsets of 13 to all ($\approx$200) training prompts per fold, and the tuned readouts with seeds pooled on subsets of 25, 50 and 100, and evaluate on the same test folds.

\subsection{Early warning}
\label{sec:earlymethod}
Can a sample that will float be caught partway through sampling and replaced? A sample fails if its final motion has any floating. We compare four strategies and count cost in denoising steps per delivered motion (1 = one full sample): generating once; \emph{finish, then filter}, which checks the final motion and regenerates if it floats; \emph{early abort}, which scores the sample at 20\% or 40\% of denoising and restarts it if the score is high; and early abort followed by the final check. Retries stop after a budget of attempts, and the last attempt is always delivered. The early score is (i) a training-free rule, abort if the decoded estimate already shows floating; (ii) a ridge readout, trained out of fold and set to abort the same share of samples, of everything the estimate shows (its features averaged over frames and its per-sample failure rates), of the frame-averaged activations of one layer (layer 3 of the 50-step model and layer 6 of the 1000-step model, each chosen on the first seed), or of both; (iii) random aborts at the same rate; or (iv) a perfect detector, as a ceiling. All strategies are simulated with the same random draws, on the 6,000 recorded MDM samples. Each already contains its estimate at every recorded step and its final motion, so every decision uses what an online run would see; a retry draws another recorded sample of the same prompt (200 simulated requests per prompt). Strategies that end with the final check are compared at equal delivered failure by varying the budget from 1 to 20 attempts.

\section{Results}
\label{sec:results}

\subsection{What goes wrong physically}
\label{sec:physics}

\begin{table}[t]
\centering
\caption{Physics of the generated motions (range over seeds; 1,000 motions per seed). Jump gravity is the median fitted fall rate of unsupported airborne stretches that do fall (real flight: 9.81\,m/s$^2$); floating counts unsupported stretches that barely fall. Here ``jump prompts'' are the 14 curated jump prompts of the main runs (Section~\ref{sec:models}); elsewhere the term means the jump-heavy runs.}
\label{tab:physics}
\begin{tabular}{lcc}
\toprule
Measure & 50-step (3 seeds) & 1000-step (3 seeds) \\
\midrule
Floating, \% of frames, jump prompts & 0.6--0.8 & 4.8--9.3 \\
Floating, \% of frames, test captions & 3.8--4.7 & 8.6--9.7 \\
Jump gravity, median (m/s$^2$), jump prompts & 9.54--9.75 & 9.13--9.22 \\
Jump gravity, median (m/s$^2$), test captions & 9.07--9.84 & 8.59--9.78 \\
Raised contact ($>$5\,cm), \% of contact frames, jump prompts & 1.2--2.4 & 3.0--8.4 \\
Raised contact ($>$5\,cm), \% of contact frames, test captions & 2.9--3.3 & 4.9--5.4 \\
\bottomrule
\end{tabular}

\end{table}

When MDM's bodies leave the ground and fall, they fall at close to the right rate (Table~\ref{tab:physics}): the median fitted gravity on the curated jump prompts is 9.54 to 9.75\,m/s$^2$ for the 50-step model and 9.13 to 9.22\,m/s$^2$ for the 1000-step model, somewhat slower than real jumps in the processed form of its training data (median 10.33\,m/s$^2$, Appendix~\ref{app:realmotion}). Individual stretches spread widely around these medians (standard deviation 2.24\,m/s$^2$ on the jump-heavy runs, against 0.7 to 1.0\,m/s$^2$ for real jumps). The common failure is floating. On HumanML3D test captions, 3.8 to 4.7\% of the 50-step model's frames lie in floating stretches, against 8.6 to 9.7\% for the 1000-step model. On the curated jump prompts the figures are 0.6 to 0.8\% and 4.8 to 9.3\%. The raised-contact rate, a foot in contact but more than 5\,cm up, follows the same pattern. Part of the floating is implied by the prompt: a person ``walking on a balance beam'' is generated without a beam and floats in nearly every sample. On test captions the original 1000-step checkpoint floats about twice as often as the 50-step one, in all three seeds and under every threshold setting we tried (Appendix~\ref{app:sensitivity}). On real motion capture the same labels flag floating in 0.09\% of frames (Appendix~\ref{app:realmotion}).

A typical floating stretch is an ordinary walking pose with the whole body several centimeters off the floor for about half a second, in a motion that touches the floor elsewhere (Figure~\ref{fig:teaser}, Appendix~\ref{app:labelfix}). Our first definition of airborne looked only at the feet, so it fitted such stretches, and stretches supported by the hands or knees, as flight. That made the 1000-step model's jumps appear to fall far too slowly (medians of 3.75 and 7.55\,m/s$^2$ in the two seeds labelled that way), an effect that disappears once floating is separated out.

\subsection{Physics is readable early in denoising}
\label{sec:readable}

\begin{figure}[t]
\centering
\includegraphics[width=\linewidth]{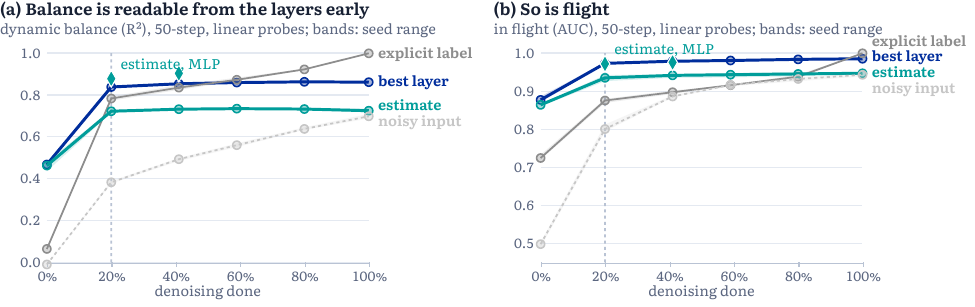}
\caption{Probe scores across denoising for the 50-step model (lines: mean over three seeds; bands: range). The best layer (deep blue) reads the final motion's dynamic balance and flight early, and so does an MLP reading everything the model's own estimate shows (teal diamonds, measured at 20\% and 40\%; bars: standard deviation). In most figures, teal marks readouts of the estimate, deep blue the model's layers, gray the context, and red the physics of a motion (airborne arcs, floating).}
\label{fig:steps}
\end{figure}

\begin{table}[t]
\centering
\caption{The standard comparison at 20\% of denoising (mean $\pm$ standard deviation over seeds). Best layer: the best of the eight layers per seed, linear probe. Explicit: the label computed from the model's clean-motion estimate. Estimate, linear: a linear probe on the estimate's 263 features. Noisy input: a linear probe on $\vx_t$.}
\label{tab:naive}
\begin{tabular}{llcccc}
\toprule
Model & Target & Best layer & Explicit & Estimate, linear & Noisy input \\
\midrule
\multirow{5}{*}{50-step} & Dynamic balance ($R^2$) & 0.838 $\pm$ 0.001 & 0.783 $\pm$ 0.009 & 0.723 $\pm$ 0.007 & 0.383 $\pm$ 0.006 \\
 & Static balance ($R^2$) & 0.759 $\pm$ 0.014 & 0.769 $\pm$ 0.019 & 0.673 $\pm$ 0.014 & 0.241 $\pm$ 0.007 \\
 & Lowest foot ($R^2$) & 0.868 $\pm$ 0.022 & 0.921 $\pm$ 0.005 & 0.817 $\pm$ 0.029 & 0.396 $\pm$ 0.022 \\
 & In flight (AUC) & 0.973 $\pm$ 0.002 & 0.876 $\pm$ 0.005 & 0.935 $\pm$ 0.008 & 0.801 $\pm$ 0.010 \\
 & Flight gravity ($R^2$) & 0.820 $\pm$ 0.035 & 0.707 $\pm$ 0.079 & 0.547 $\pm$ 0.053 & 0.209 $\pm$ 0.046 \\
\midrule
\multirow{5}{*}{1000-step} & Dynamic balance ($R^2$) & 0.697 $\pm$ 0.029 & 0.655 $\pm$ 0.008 & 0.550 $\pm$ 0.031 & 0.175 $\pm$ 0.003 \\
 & Static balance ($R^2$) & 0.560 $\pm$ 0.058 & 0.546 $\pm$ 0.019 & 0.508 $\pm$ 0.066 & 0.181 $\pm$ 0.009 \\
 & Lowest foot ($R^2$) & 0.804 $\pm$ 0.020 & 0.860 $\pm$ 0.008 & 0.766 $\pm$ 0.012 & 0.303 $\pm$ 0.016 \\
 & In flight (AUC) & 0.926 $\pm$ 0.003 & 0.788 $\pm$ 0.003 & 0.893 $\pm$ 0.001 & 0.759 $\pm$ 0.010 \\
 & Flight gravity ($R^2$) & 0.692 $\pm$ 0.023 & 0.691 $\pm$ 0.060 & 0.405 $\pm$ 0.176 & 0.193 $\pm$ 0.022 \\
\bottomrule
\end{tabular}

\end{table}

Linear probes recover the final motion's physics from the activations well before sampling ends (Figure~\ref{fig:steps}, Table~\ref{tab:naive}). At 20\% of denoising, the best layer of the 50-step model reaches $R^2 = 0.838 \pm 0.001$ for dynamic balance, 0.868 for the lowest-foot height and 0.820 for flight gravity, and an AUC of 0.973 for flight. In these data the flight-gravity score mostly reflects whether an airborne body floats or falls (Section~\ref{sec:fairresults}). Layer 7 is the best layer for dynamic balance in all six runs. The result transfers across prompt types: for dynamic balance at 20\%, the best layer scores 0.835 on curated prompts and 0.839 on test captions. The 1000-step model is less readable at the same fraction of denoising (0.697 for dynamic balance).

Taken at face value, the standard comparison suggests hidden knowledge. For dynamic balance, flight and, on the 50-step model, gravity, the best layer beats both the explicit label and the linear probe on the estimate at the same step. For dynamic balance the gap over the explicit label is $+0.045$ to $+0.065$ in the three 50-step seeds and $+0.018$ to $+0.060$ in the three 1000-step seeds. Only the lowest-foot height, which the estimate shows directly, and static balance on the 50-step model (0.769 against 0.759) favor the explicit label, and flight gravity on the 1000-step model is a tie (0.692 against 0.691).

\subsection{The advantage disappears under a fair readout}
\label{sec:fairresults}

\begin{table}[t]
\centering
\caption{Tuned readouts at 20\% of denoising, seeds pooled, every readout tuned on held-out prompts (Section~\ref{sec:fair}). Pooled $R^2$, or AUC for flight and floats or falls; the best tuned readout in bold. Both: estimate and layer, tuned with fallback. Gain: Both minus Estimate, with 95\% prompt-bootstrap intervals, pooled and within prompt and frame. ``Jump prompts'': the jump-heavy runs. --: too few prompts for an interval. 0 (all folds): the fallback kept the estimate in every fold, so the gain is zero by construction (Table~\ref{tab:gaindetail}).}
\label{tab:tuned}
\resizebox{\linewidth}{!}{\begin{tabular}{llccccccc}
\toprule
& & & & \multicolumn{3}{c}{Tuned on held-out prompts} & \multicolumn{2}{c}{Gain from adding the layer [95\% CI]} \\
\cmidrule(lr){5-7} \cmidrule(lr){8-9}
Target & Model & Explicit & Layer, ridge & Estimate & Layer & Both & Pooled score & Within prompt and frame \\
\midrule
Dynamic balance & 50-step & 0.784 & 0.850 & \textbf{0.889} & 0.863 & \textbf{0.889} & $+0.000$ [$-0.001$, $+0.001$] & $+0.000$ [$-0.001$, $+0.001$] \\
Dynamic balance & 1000-step & 0.655 & 0.720 & \textbf{0.779} & 0.765 & 0.777 & $-0.002$ [$-0.005$, $-0.000$] & $-0.001$ [$-0.003$, $+0.000$] \\
Dynamic balance & MotionDiffuse & 0.124 & 0.620 & \textbf{0.823} & 0.700 & \textbf{0.823} & $0$ (all folds) & $0$ (all folds) \\
\addlinespace[2pt]
Static balance & 50-step & 0.771 & 0.780 & \textbf{0.830} & 0.800 & 0.828 & $-0.001$ [$-0.004$, $+0.002$] & $-0.002$ [$-0.006$, $+0.002$] \\
Static balance & 1000-step & 0.548 & 0.592 & \textbf{0.671} & 0.647 & \textbf{0.671} & $0$ (all folds) & $0$ (all folds) \\
Static balance & MotionDiffuse & 0.103 & 0.537 & \textbf{0.729} & 0.547 & \textbf{0.729} & $0$ (all folds) & $0$ (all folds) \\
\addlinespace[2pt]
In flight & 50-step & 0.876 & 0.977 & \textbf{0.976} & 0.971 & 0.972 & $-0.004$ [$-0.008$, $-0.000$] & $-0.008$ [$-0.014$, $-0.002$] \\
In flight & 1000-step & 0.788 & 0.934 & \textbf{0.935} & 0.932 & \textbf{0.935} & $-0.000$ [$-0.003$, $+0.002$] & $+0.002$ [$-0.000$, $+0.005$] \\
In flight & MotionDiffuse & 0.778 & 0.951 & \textbf{0.943} & 0.905 & 0.939 & $-0.004$ [$-0.009$, $-0.001$] & $-0.001$ [$-0.006$, $+0.003$] \\
In flight & 50-step, jump prompts & 0.926 & 0.981 & \textbf{0.984} & 0.981 & 0.983 & $-0.001$ [$-0.002$, $-0.000$] & $-0.001$ [$-0.002$, $+0.001$] \\
In flight & 1000-step, jump prompts & 0.878 & 0.955 & 0.960 & 0.961 & \textbf{0.962} & $+0.001$ [$-0.000$, $+0.003$] & $+0.001$ [$-0.002$, $+0.003$] \\
In flight & MotionDiffuse, jump prompts & 0.734 & 0.978 & \textbf{0.975} & 0.972 & \textbf{0.975} & $+0.000$ [$-0.001$, $+0.002$] & $+0.002$ [$-0.000$, $+0.004$] \\
\addlinespace[2pt]
Floats or falls & 50-step & 0.931 & 0.990 & 0.980 & \textbf{0.988} & \textbf{0.988} & $+0.008$ [$+0.001$, $+0.015$] & $+0.050$ [$-0.007$, $+0.099$] \\
Floats or falls & 1000-step & 0.922 & 0.980 & 0.964 & \textbf{0.976} & 0.967 & $+0.003$ [$-0.001$, $+0.008$] & $+0.008$ [$-0.013$, $+0.029$] \\
Floats or falls & MotionDiffuse & 0.883 & 0.988 & 0.940 & \textbf{0.988} & 0.966 & $+0.026$ [$+0.007$, $+0.055$] & -- \\
Floats or falls & 50-step, jump prompts & 0.837 & 0.971 & \textbf{0.974} & 0.969 & \textbf{0.974} & $-0.000$ [$-0.009$, $+0.008$] & $-0.000$ [$-0.014$, $+0.016$] \\
Floats or falls & 1000-step, jump prompts & 0.667 & 0.973 & 0.966 & \textbf{0.969} & 0.967 & $+0.001$ [$-0.001$, $+0.003$] & $-0.002$ [$-0.009$, $+0.006$] \\
Floats or falls & MotionDiffuse, jump prompts & 0.741 & 0.836 & 0.827 & 0.784 & \textbf{0.828} & $+0.001$ [$-0.017$, $+0.017$] & $+0.004$ [$-0.018$, $+0.024$] \\
\addlinespace[2pt]
Fall rate (falling only) & 50-step & 0.171 & 0.267 & $-0.168$ & \textbf{0.217} & 0.084 & $+0.253$ [$+0.060$, $+0.505$] & $+0.046$ [$-0.041$, $+0.127$] \\
Fall rate (falling only) & 1000-step & 0.209 & 0.115 & $-0.069$ & \textbf{0.090} & 0.013 & $+0.082$ [$-0.035$, $+0.167$] & $+0.106$ [$+0.020$, $+0.195$] \\
Fall rate (falling only) & 50-step, jump prompts & 0.152 & 0.374 & 0.351 & 0.427 & \textbf{0.428} & $+0.077$ [$+0.051$, $+0.107$] & $+0.073$ [$+0.045$, $+0.098$] \\
Fall rate (falling only) & 1000-step, jump prompts & 0.161 & 0.273 & 0.287 & 0.309 & \textbf{0.363} & $+0.076$ [$+0.053$, $+0.096$] & $+0.021$ [$+0.006$, $+0.037$] \\
Fall rate (falling only) & MotionDiffuse, jump prompts & 0.289 & 0.326 & 0.365 & 0.326 & \textbf{0.390} & $+0.025$ [$+0.003$, $+0.047$] & $+0.017$ [$-0.001$, $+0.036$] \\
\bottomrule
\end{tabular}
}
\end{table}

\begin{figure}[t]
\centering
\includegraphics[width=\linewidth]{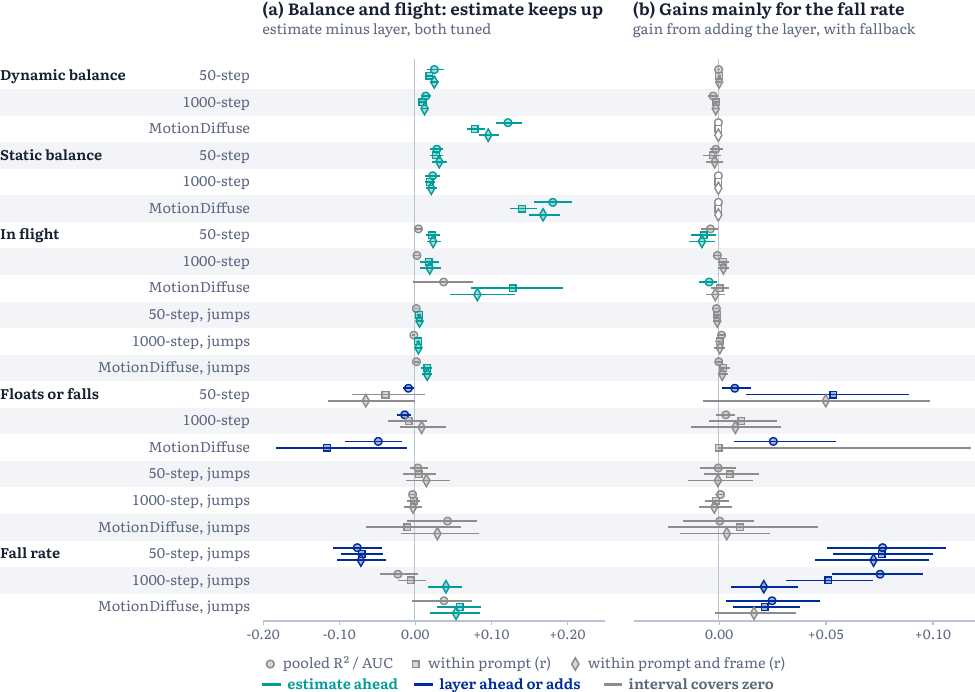}
\caption{The fair comparison (tuned readouts, seeds pooled, 95\% prompt-bootstrap intervals). (a)~Tuned estimate minus tuned layer: positive favors the estimate. (b)~The gain from adding the layer to the estimate, with fallback. Marker shape gives the score; color gives the verdict: teal where the estimate is ahead, deep blue where the layer is ahead or adds, gray where the interval covers zero, and open markers at zero where the fallback kept the estimate in every fold. In (b), clear gains appear only for the fall rate on the jump prompts (``jumps'') and for floating in the small main data sets; balance and flight show none. Not drawn: scores without an interval (too few prompts) and the main data's fall rate (1,147 and 1,620 falling frames, intervals up to about $\pm 0.25$; Table~\ref{tab:tuned}).}
\label{fig:gain}
\end{figure}

\begin{figure}[t]
\centering
\includegraphics[width=\linewidth]{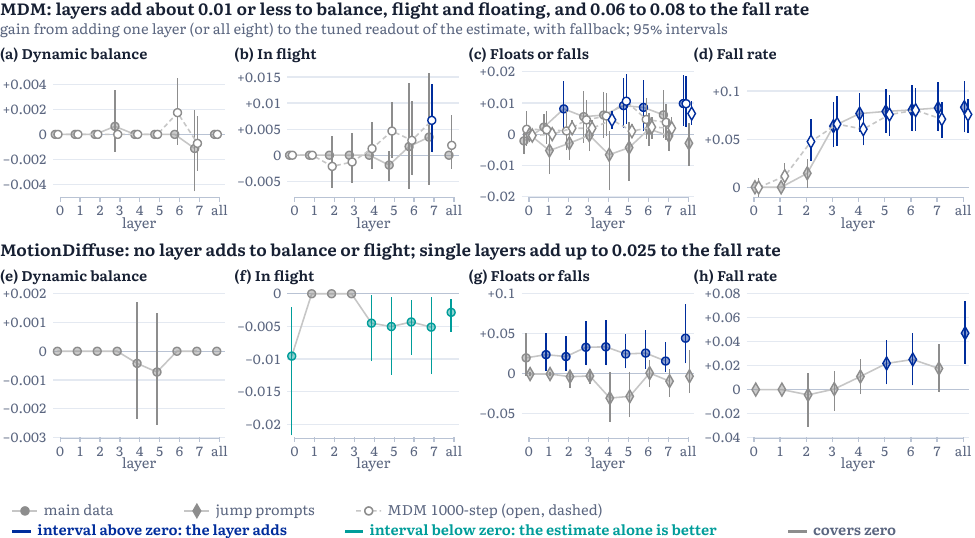}
\caption{The gain from adding one layer, or all eight, to the tuned readout of the estimate (with fallback, 20\% of denoising, 95\% intervals; colors as in Figure~\ref{fig:gain}). Circles: main data; diamonds: jump prompts; open markers and dashed lines: the 1000-step model. Top row, MDM: only the fall rate rises well above zero, from layer~3 on (layer~2 on the 1000-step model); the floating gains stay at about 0.01 or less. Bottom row, MotionDiffuse: the same pattern with a smaller fall-rate gain, from layers 5 to 7, and a floating gain only on the main data's 50 prompts. The main data's fall rate is not shown (too few falling frames).}
\label{fig:layers}
\end{figure}

Table~\ref{tab:tuned} and Figure~\ref{fig:gain} compare the readouts with the three seeds of each model pooled and every readout tuned on held-out prompts (Section~\ref{sec:fair}). A first version with one fixed MLP per seed gives the same ordering (Appendix~\ref{app:fixedmlp}). For balance and flight the estimate is read as well as the layer: for both balance measures its tuned readout beats the tuned layer, by 0.015 to 0.029 in $R^2$ on both MDM checkpoints and by 0.12 and 0.18 on MotionDiffuse, and for flight the estimate ties or leads. Adding the layer to the estimate gains nothing. With the readout that may fall back to the estimate, every interval covers zero or lies below it, across prompts and within prompt and frame, and so does every interval in two refits with other validation splits and initializations, run for dynamic balance on the 50-step model. The layer sweep shows the same for every layer: none adds more than 0.007, and only one interval, for flight at layer~7 of the 1000-step model, barely excludes zero (Figure~\ref{fig:layers}). On MDM, whose conditional estimate is a fixed linear map of layer~7, a gain from that layer could come only from directions the output head discards.

This null result is not produced by the estimator. Without the fallback, the larger readout scores within 0.003 of the estimate or below it for balance and flight, so the fallback hides no gain there. It keeps the estimate alone in one to five of the five folds (Table~\ref{tab:gaindetail}). On the scale of Proposition~\ref{prop:vinfo}, the largest balance gain is 0.001 nats, against 0.063 for the fall rate below.

A positive control shows what the protocol can detect (Table~\ref{tab:control}). We take the estimate readout's out-of-fold residual, add noise so that its best linear use would raise $R^2$ by 0.005, 0.01 or 0.02 on the squared-error scale, add it to the layer's activations along a random direction, and rerun the protocol unchanged on the 50-step model. For dynamic balance even the smallest signal is found: the gain is $+0.013$ [$0.009$, $0.017$], more than the linear amount because the tuned readouts extract more, against $+0.000$ [$-0.001$, $0.001$] for the real layer. For flight and floating the pooled AUC, near 0.97, barely moves even at 0.02, partly because the readouts are selected by squared error. The within-prompt-and-frame score does find the signal, from 0.005 for flight and from 0.01 for floating on the jump prompts, and on that score the real layer adds nothing ($-0.008$ and $0.000$). For these two targets the evidence for no gain is therefore that score, and a gain below 0.01 for floating could go undetected.

Two details show where the estimate's advantage comes from. First, the readout family matters: restricted to linear readouts, the layer does add a little (up to 0.02 on MDM, 0.05 to 0.07 on MotionDiffuse), so part of what the estimate shows needs a nonlinear readout. Second, on MDM the input matters more than the family: from its 263 raw features alone the tuned estimate only ties the tuned layer, and the physics labels computed from it add about 0.02 for dynamic balance (Table~\ref{tab:raw}). On MotionDiffuse the raw features alone already beat the tuned layer, by 0.12 and 0.18 (Table~\ref{tab:tuned}), and the labels add nothing. A temporal readout that sees all 30 recorded frames gives the same ordering in all six runs (Appendix~\ref{app:impl}). In the terms of Section~\ref{sec:fair}, the usable information the layer adds about balance and flight depends on the readout family, and for the most flexible readouts we tried it is indistinguishable from zero.

\paragraph{Flight gravity: floating and falling.} The fitted flight gravity mixes two questions. In the main data most of its labelled frames (76\% and 83\%) lie in floating stretches, so its score mostly measures whether an airborne body floats or falls, and only 57 and 78 prompts contain falling frames, too few for reliable tuned readouts. We therefore split the target (Section~\ref{sec:labels}) and use the jump-heavy runs, in which most labelled frames fall (11,640 and 13,444 frames from 233 and 240 prompts).

Whether an airborne body floats or falls is visible in the estimate on the jump prompts, where 236 prompts have airborne stretches on the 50-step model. Its tuned readout reaches an AUC of 0.97 on both MDM checkpoints, and no single layer, nor all eight together, adds more than 0.007 in pooled AUC (Figure~\ref{fig:layers}). Across the three fits the layer adds at most $0.004$ in pooled AUC, and $0.019$ [$0.008$, $0.030$] within prompt in one of them. The layer adds more only in the smaller main data, where just 106 prompts have airborne stretches. There, on the 50-step model, it adds $+0.008$ [$0.001$, $0.015$] in pooled AUC and $+0.054$ [$0.013$, $0.089$] in the within-prompt correlation, and in two refits up to $0.014$ pooled and $0.08$ to $0.11$ within prompt and frame.

\begin{figure}[t]
\centering
\includegraphics[width=\linewidth]{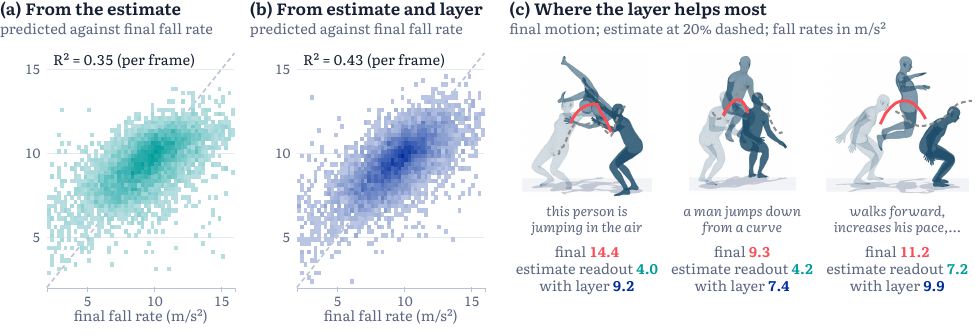}
\caption{Held-out predictions of the final fall rate on the jump prompts (50-step model, 20\% of denoising, tuned readouts; density of falling stretches in cells of 0.25\,m/s$^2$, darker for more). (a)~From the estimate alone and (b)~with layer~5 added (with fallback); dashed: perfect prediction; $R^2$ per frame, as in Table~\ref{tab:tuned}. (c)~Among jumps that travel at least 0.4\,m, the three where adding the layer reduces the error most, selected for that reason: the final motion at take-off, apex and landing, its center-of-mass path (red while airborne), and the estimate's path at 20\% (dashed).}
\label{fig:layerknows}
\end{figure}

\begin{figure}[t]
\centering
\includegraphics[width=\linewidth]{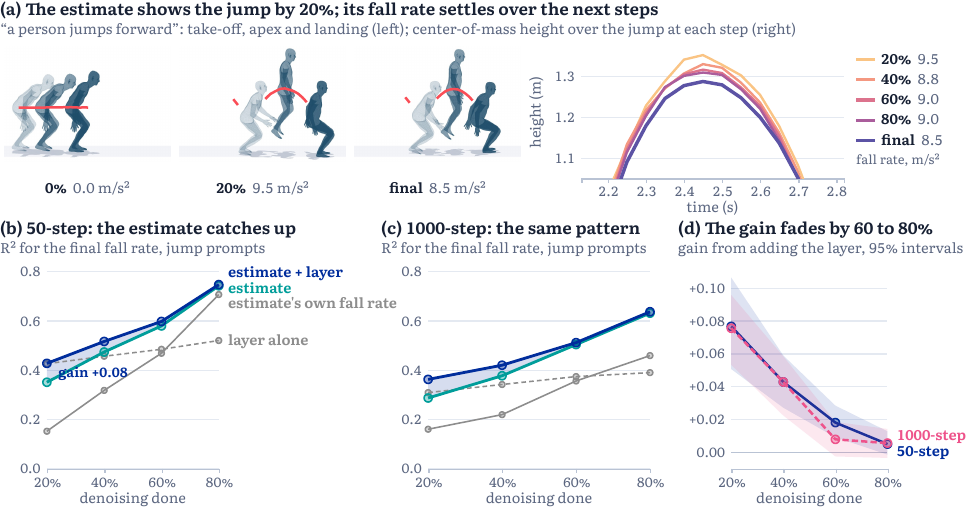}
\caption{The fall rate of jumps across denoising (jump prompts, layer~5, tuned readouts, seeds pooled). (a)~A typical jump, drawn from the estimate at 0\% and 20\% of denoising and from the final motion (center of mass red while airborne), and its center-of-mass height at every recorded step with that step's fitted fall rate. At 0\% the estimate is a crouched average pose; from 20\% the jump is in place, and its height and fall rate change by only a few centimeters and about 1\,m/s$^2$ afterwards. The jump is chosen by a fixed rule: among falling stretches of the 50-step runs that travel at least 0.4\,m, the one whose 20\% fall-rate error is the median (1.0\,m/s$^2$). (b, c)~$R^2$ for the final fall rate from the estimate and from estimate and layer together; the shaded gap is the gain. Gray: the estimate's own fitted fall rate, and the layer alone. (d)~The gain, with 95\% intervals: largest at 20\%, below 0.02 by 60\%.}
\label{fig:stepsfall}
\end{figure}

The fall rate of the stretches that do fall is the one case in which the layers know more. On the jump prompts of the 50-step model, adding the layer raises $R^2$ from 0.35 to 0.43, a gain of $+0.077$ [$0.051$, $0.107$], or 0.063 nats (Tables~\ref{tab:tuned} and~\ref{tab:gaindetail}, Figure~\ref{fig:layerknows}), with similar gains within prompt and frame and in two refits. With more training prompts the gain first shrinks and then levels off from 100 prompts on (0.085, then 0.077; Figure~\ref{fig:tunedcurve}b). It replicates on the 1000-step model ($+0.076$ [$0.053$, $0.096$]), though there it is smaller within prompt and frame ($+0.021$) and with fewer training prompts (Figure~\ref{fig:tunedcurve}c), and the layer alone is no longer clearly ahead of the estimate (0.309 against 0.287, an interval that covers zero), so the two carry partly different information. On both MDM checkpoints every layer from 3 to 7 adds 0.06 to 0.08 (Figure~\ref{fig:layers}). The standard comparison overstates even this genuine case: the layer's ridge probe beats the explicit label by three times the conditional gain on the 50-step model and 1.5 times on the 1000-step model, because the fall rate fitted to a 20\% estimate predicts the final one poorly.

Four checks leave the gain in place. It is not only the layer knowing where a jump is: in 92\% of the 11,640 falling frames, from 7,125 falling stretches, the 20\% estimate is airborne too, and there the layer still adds $+0.069$. It is not a weak readout of the estimate's timing: the estimate's readout already includes its own fitted fall rate, and a temporal readout over all 30 recorded frames still leaves the layer ahead in every seed (Appendix~\ref{app:impl}). It is not the text: giving the readout the prompt's CLIP embedding, the model's own text input, does not help on held-out prompts, and the layer's gain beyond estimate and text is unchanged. And label noise is not the ceiling: on real motion capture the gravity label's spread would allow an $R^2$ of 0.81 to 0.91 (Appendix~\ref{app:realmotion}).

The gain is, however, transient (Figure~\ref{fig:stepsfall}). At 40\% of denoising layer~5 still adds 0.043 on both MDM checkpoints (Figure~\ref{fig:stepsfall}d). By 60\% no layer adds more than 0.02, and at 80\% the interval for layer~5 covers zero, because the estimate's readout improves faster than the layer's. A separate layer sweep gives up to about 0.06 at 40\%, and the same readout differs by up to 0.016 between that sweep and the step run. At a fifth of the denoising steps, then, MDM's middle and late layers carry information about how fast a jump will fall that its estimate does not yet show in any form we could read: a step-wise case of a model knowing more than it shows, of the kind the video studies report. Such a gain must vanish by the end of sampling, when the final motion is a function of the estimate, so what the measurement adds is its size and duration: about 0.08 in $R^2$, for about two fifths of the trajectory.

\paragraph{A second model.} MotionDiffuse (Section~\ref{sec:models}) predicts the noise, receives the text in every layer and uses no guidance, and it gives the same picture for balance and flight. On its main data the tuned estimate beats the tuned layer by 0.12 in $R^2$ for dynamic balance and 0.18 for static balance, more than on MDM. Adding any one of its eight layers gains at most 0.001 for dynamic balance, and its chosen layers add nothing for static balance or flight. Floating follows MDM's pattern: on the main data, where only 50 prompts have airborne stretches, its layers add 0.016 to 0.034 in AUC and the layer alone beats the estimate (0.988 against 0.940), while on its jump prompts (2,940 motions, 235 prompts with such stretches) they add nothing, and the same holds for flight and for the unsplit flight gravity (at most $+0.002$).

The fall rate again shows a gain, but a smaller and less stable one. At 20\% of denoising (step 200 of 1000), adding layer~6 raises $R^2$ from 0.365 to 0.390, $+0.025$ [$0.003$, $0.047$] (Table~\ref{tab:tuned}, Figures~\ref{fig:gain} and \ref{fig:layers}). It holds within prompt ($+0.022$ [$0.007$, $0.038$]), but within prompt and frame its interval covers zero ($+0.017$ [$-0.001$, $0.036$]), and in two refits with other validation splits and initializations it is $+0.022$ [$0.007$, $0.041$] and 0, the fallback keeping the estimate in every fold of the second. It again comes from the later layers (layers 5 to 7 add 0.018 to 0.025 and all eight together 0.047, layers 0 to 3 nothing), the text does not explain it (estimate, text and layer together gain the same $+0.025$ over the estimate as without the text), and it fades with denoising: $+0.012$ [$0.003$, $0.022$] at 40\% and $+0.007$ [$-0.002$, $0.016$] at 60\%.

At matched noise the gap to MDM is larger still, because the same fraction of denoising is not the same noise level on the two models. At 20\% MotionDiffuse's signal-to-noise ratio is 0.002, against MDM's 0.10, and it reaches MDM's range only between 40 and 60\% (Table~\ref{tab:snr}). At 60\%, with a ratio of 0.24, between MDM's at 20\% (0.10) and 40\% (0.52), MotionDiffuse's gain is $+0.007$, against MDM's $+0.077$ and $+0.043$ at those steps.

Across targets and on all three models, then, a readout of what the model shows, tuned as carefully as the probe on its activations, accounts for what the activations predict about the final motion's balance and flight, and about floating on the jump prompts. In the smaller main data sets adding a layer does help with floating (the chosen layer gains up to 0.026 in AUC, removing up to 43\% of the misranked pairs, and single MotionDiffuse layers up to 0.034; Table~\ref{tab:gaindetail}). The apparent knowledge in Table~\ref{tab:naive} came from comparing a trained probe on the activations with an untrained or linear readout of the estimate. The fall rate of jumps is the exception, and it is found by the same controls that remove the other advantages.

\subsection{Small training sets can reverse the comparison}
\label{sec:learning}

\begin{figure}[t]
\centering
\includegraphics[width=\linewidth]{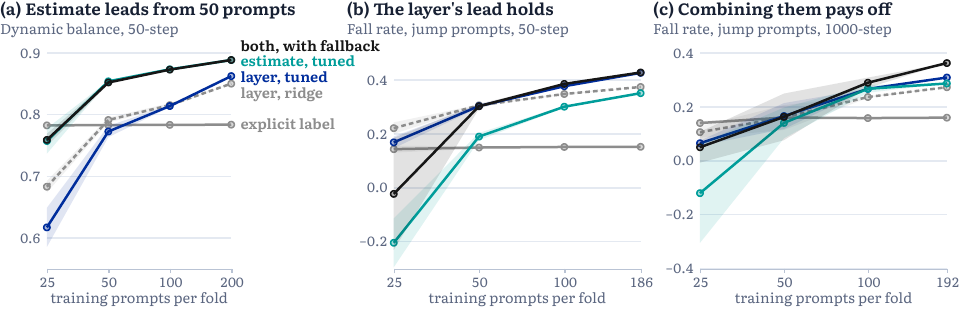}
\caption{Tuned readouts trained on fewer prompts (seeds pooled, same test folds; two draws per size below the full set; lines: mean, bands: range). (a)~Dynamic balance: the explicit label wins at 25 prompts, and from 50 on the estimate leads the layer. (b)~The fall rate of jumps, 50-step model: the layer leads at every size. (c)~The same target, 1000-step model: the gain from combining estimate and layer grows from 0.02 to 0.08 only with the full training set.}
\label{fig:tunedcurve}
\end{figure}

The pilot study pointed the other way: with 64 prompts and one held-out set of 12, the explicit label beat the best layer on dynamic balance (0.832 against 0.814). Training size explains it: the one-feature explicit label barely depends on the number of training prompts, while the 512-feature linear layer probe overtakes it only between 52 and 100 training prompts (Appendix~\ref{app:fixedmlp}, Figure~\ref{fig:learning}), and the pilot's 12-prompt test set was also too small to fix the sign. Tuned readouts with pooled seeds show the same dependence (Figure~\ref{fig:tunedcurve}). For dynamic balance, the one-feature explicit label still beats every trained readout at 25 training prompts on the 50-step model (0.783, against 0.757 for the tuned estimate and 0.617 for the tuned layer), and from 50 prompts on the tuned estimate leads the tuned layer at every size (0.854 against 0.773 at 50, 0.889 against 0.863 at 200). The lead narrows as data grows, from 0.060 at 100 prompts to 0.026 at 200 (0.026 to 0.015 on the 1000-step model), so with much larger training sets the layer's readout could catch up. The gain from adding the layer, which is what bears on hidden knowledge, stays at zero at every size we measured ($-0.003$ to $+0.003$ with the fallback readout). For the fall rate of jumps on the 50-step model the layer leads at every size, and its lead over the estimate stops shrinking: 0.113 at 50 prompts, 0.076 at 100 and 0.075 at 186. On the 1000-step model the layer alone leads by at most 0.026 from 50 prompts on, and the gain from adding it to the estimate is 0.022 and 0.023 at 50 and 100 prompts and 0.076 at 192. A comparison at one training size can favor either side, and only a lead that persists as data grows is evidence about the model. Video probing benchmarks are of a similar size: the IntPhys development set used in Section~\ref{sec:video} has 360 videos.

\subsection{Early warning: the estimate is enough, and the payoff is small}
\label{sec:early}

\begin{table}[t]
\centering
\begin{minipage}[t]{0.37\linewidth}
\centering
\caption{The training-free rule ``abort if the decoded estimate shows any floating''. Precision: flagged samples that do float. Recall: floating samples flagged.}
\label{tab:detector}
\resizebox{\linewidth}{!}{\begin{tabular}{llccc}
\toprule
Model & Check at & Flagged & Precision & Recall \\
\midrule
50-step & 20\% & 7.0\% & 80\% & 59\% \\
50-step & 40\% & 7.0\% & 88\% & 65\% \\
1000-step & 20\% & 10.1\% & 82\% & 47\% \\
1000-step & 40\% & 11.1\% & 84\% & 52\% \\
\bottomrule
\end{tabular}
}
\end{minipage}\hfill
\begin{minipage}[t]{0.6\linewidth}
\centering
\caption{Compute saved by aborting at 20\% of denoising, with the final check kept, against finish-then-filter at the same delivered failure (its budget of 2, 3 or 5 tries). The trained readouts (ridge on frame-averaged inputs: everything the estimate shows, one layer, or both) abort the same share of samples as the rule.}
\label{tab:savings}
\resizebox{\linewidth}{!}{\begin{tabular}{lccccccc}
\toprule
Model & Delivered & Finish, then filter & \multicolumn{5}{c}{Compute saved by early abort} \\
\cmidrule(lr){4-8} & floating & (compute) & estimate rule & estimate readout & layer probe & readout + layer & perfect detector \\
\midrule
50-step & 3.8\% & 1.24 & 6.7\% & 6.8\% & 4.5\% & 4.5\% & 14.9\% \\
50-step & 4.5\% & 1.15 & 4.8\% & 5.1\% & 5.1\% & 5.5\% & 10.2\% \\
50-step & 5.5\% & 1.09 & 3.4\% & 3.6\% & 3.6\% & 3.8\% & 6.7\% \\
1000-step & 4.6\% & 1.39 & 9.9\% & 9.8\% & 10.7\% & 10.2\% & 22.5\% \\
1000-step & 6.8\% & 1.27 & 6.9\% & 6.9\% & 6.8\% & 6.5\% & 17.0\% \\
1000-step & 9.4\% & 1.18 & 4.3\% & 4.4\% & 4.0\% & 3.7\% & 11.9\% \\
\bottomrule
\end{tabular}
}
\end{minipage}
\end{table}

\begin{figure}[t]
\centering
\includegraphics[width=\linewidth]{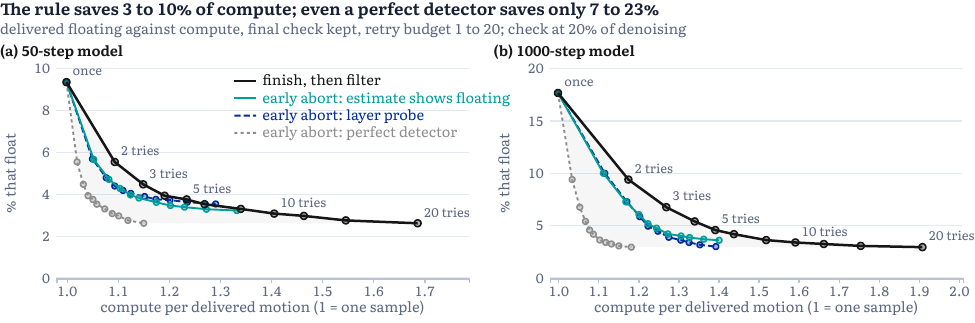}
\caption{Delivered floating against compute, with the final check kept and the retry budget varied from 1 to 20 tries (labels on the finish-then-filter curve). Aborting at 20\% of denoising when the decoded estimate shows floating (teal) moves the curve only slightly, and a layer probe (deep blue) moves it about as much. The shaded region between finish-then-filter and a perfect detector bounds what any early signal could save.}
\label{fig:early}
\end{figure}

An early signal is useful in practice if it stops samples that will fail. For MDM, the model's own estimate provides one without any training (Table~\ref{tab:detector}). If the estimate shows floating at 20\% of denoising, the final motion floats in 80\% (50-step) and 82\% (1000-step) of cases, and the rule catches 59\% and 47\% of all floating samples. At 40\%, precision rises to 88\% and 84\%. As a per-sample score, the estimate's own floating rate also ranks samples within a prompt better than trained readouts. At 20\%, its Spearman correlation with the final floating rate after removing prompt means is 0.57 (50-step) and 0.46 (1000-step), against 0.34 and 0.33 for a ridge readout of everything the estimate shows, 0.26 and 0.28 for the same ridge on a layer, and 0.29 and 0.28 for the two together: at the sample level, too, the layer adds nothing to the estimate's readout, at least for these ridge readouts of frame-averaged inputs. The prompt's text embedding alone predicts the floating rate only weakly on held-out prompts (Spearman 0.25 and 0.23).

The compute payoff is small (Table~\ref{tab:savings}, Figure~\ref{fig:early}). Checking the final motion costs almost nothing, so finish-then-filter is the baseline to beat: with two tries it delivers 5.5\% floating motions at 1.09 times the compute of a single sample (50-step), and 9.4\% at 1.18 (1000-step), against 9.4\% and 17.7\% when generating once. Early abort without the final check bottoms out at 6.6 to 6.9\% and 13.6 to 14.5\% floating with the training-free rule, and at 5.5 to 6.3\% and 10.3 to 12.0\% with the trained readouts. Without the final check, the rule at 20\% of denoising delivers within 0.9 percentage points of finish-then-filter at the same compute (mixing budgets of one and two tries), and at 40\% it falls behind. With the final check kept, aborting at 20\% with the rule saves 3.4 to 6.7\% of compute at matched failure for the 50-step model and 4.3 to 9.9\% for the 1000-step model, more at stricter targets. The trained readouts, set to abort the same share of samples, save about as much (3.6 to 10.7\%) with no consistent order between them: the readout of the estimate, the layer probe, and the two together. At the strictest 50-step target the two that use the layer save less (4.5\%, against 6.7 to 6.8\%). Random aborts give no gain. Even a perfect detector would save only 6.7 to 14.9\% and 11.9 to 22.5\%. This ceiling has a closed form.
\begin{proposition}\label{prop:ceiling}
Let prompt $u$ fail in a share $p_u$ of its samples, let an attempt aborted early cost a fraction $\tau$ of a full sample, and compare strategies at the same retry budget $B$ with the final check kept, prompts weighted equally. A perfect detector saves the fraction
\begin{equation}
    S_B \;=\; (1-\tau)\, \frac{\sum_u \sum_{j=1}^{B-1} p_u^{\,j}}{\sum_u \sum_{j=0}^{B-1} p_u^{\,j}}
\end{equation}
of the compute of finish-then-filter at the same delivered failure: $1-\tau$ times the share of finish-then-filter's attempts that are discarded. At every budget, no other detector that checks at the same step uses less compute or delivers fewer failures.
\end{proposition}
With the measured $p_u$ and $\tau = 0.22$ (50-step) and $0.201$ (1000-step), the formula gives 0.067, 0.101 and 0.148 at budgets 2, 3 and 5 for the 50-step model, and 0.120, 0.172 and 0.226 for the 1000-step model, within 0.0015 of the simulation (proof in Appendix~\ref{app:proofs}). The ceiling is low because floating affects only 9 to 18\% of samples and concentrates in particular prompts: after 20 tries, 2.6 to 3.7\% of delivered motions still float, from prompts such as walking on a balance beam that float in nearly every sample.

\subsection{A related readout issue in a video diffusion model}
\label{sec:video}

\begin{table}[t]
\centering
\caption{IntPhys development set, plausibility read from the clean Wan~2.1 VAE latent (360 videos, 90 scenarios; scenario-grouped 5-fold cross-validation, 10 repeats, 3 for the nonlinear temporal probe; mean $\pm$ standard deviation). Pairwise: share of possible/impossible pairs within a scenario in which the possible video scores higher. Chance is 50\% for both.}
\label{tab:intphys}
\begin{tabular}{lcc}
\toprule
Readout of the clean VAE latent & Per-video (\%) & Pairwise (\%) \\
\midrule
\multicolumn{3}{l}{\emph{Linear in the latent}} \\
\quad Pooled over time and space & 50.1 $\pm$ 0.5 & 52.1 $\pm$ 1.5 \\
\quad Time kept & 51.5 $\pm$ 0.7 & 50.4 $\pm$ 1.3 \\
\quad Time + $4{\times}4$ space & 50.9 $\pm$ 1.1 & 54.4 $\pm$ 1.2 \\
\quad Time + $4{\times}4$ space + signed changes & 52.8 $\pm$ 1.1 & 52.2 $\pm$ 2.2 \\
\multicolumn{3}{l}{\emph{Linear after a fixed nonlinearity: absolute frame-to-frame change}} \\
\quad Time kept + $|\Delta|$ & 53.4 $\pm$ 1.2 & 57.4 $\pm$ 2.2 \\
\quad $4{\times}4$ space, $|\Delta|$ only & 61.8 $\pm$ 0.6 & 77.9 $\pm$ 1.7 \\
\quad Time + $4{\times}4$ space + $|\Delta|$ & 62.1 $\pm$ 0.9 & 79.6 $\pm$ 2.3 \\
\multicolumn{3}{l}{\emph{Trained nonlinear readout}} \\
\quad Small temporal CNN & 50.4 $\pm$ 0.3 & 53.9 $\pm$ 2.5 \\
\midrule
Shuffled labels (1,000 shuffles, last $|\Delta|$ row; mean, 95th pct.) & 50.1, 53.1 & 49.9, 56.7 \\
\bottomrule
\end{tabular}

\end{table}

\citet{esmati2026invisible} decode plausibility linearly from the transformer states of video diffusion models, including the 1.3B Wan model \citep{wan2025}, find the clean VAE latent at chance, and conclude that the diffusion model constructs the signal (Section~\ref{sec:related}). Their transformer probes act on states averaged over all tokens. Their public code does not include the latent baseline, so we reproduce it by averaging the latent in the same way. We encode the IntPhys development set \citep{riochet2018intphys} with the Wan~2.1 VAE at the resolution of their IntPhys setting ($256 \times 256$; 97 frames, giving a latent of 16 channels by 25 frames by $32 \times 32$) and read plausibility with ridge classifiers, keeping the four versions of each scenario on the same side of every split (Appendix~\ref{app:video}).

The fully averaged latent is at chance, as they report (Table~\ref{tab:intphys}), and so is every readout that is linear in the latent: keeping the time axis, alone or with a coarse $4 \times 4$ spatial grid, and adding signed frame-to-frame differences, which are themselves linear in the latent, all stay between 50 and 55\% pairwise. One fixed nonlinearity changes this. The absolute frame-to-frame change of the $4 \times 4$ grid features, read linearly, reaches 77.9\% pairwise accuracy on its own, and 79.6\% pairwise and 62.1\% per-video together with the grid features. That readout scores 79 to 82\% pairwise in each of the three IntPhys blocks, and it or the grid's absolute change alone is the best of the readouts we tried in each (Appendix~\ref{app:video} lists them all), so choosing it on any one block and testing on the other two gives the same result. It also beats a label-shuffled null (mean 49.9\%, 95th percentile 56.7\%) in all 1,000 shuffles ($p = 0.001$), and the signed changes do not ($p = 0.31$ pairwise, $0.06$ per video). Without the spatial grid the same transform reaches only 57.4\%. The signal is not a global artifact of how the impossible videos are spliced together: the mean frame-to-frame change is only 1.00 to 1.03 times larger in impossible videos at every step, and training-free ``largest change'' scores, over the whole frame or over any cell of the $4 \times 4$ grid, are at chance (50.0\% pairwise). The readout uses localized changes, consistent with the impossible event itself, an object vanishing, appearing or changing in one region. A small temporal CNN did not train on about 290 training videos.

The averaged baseline in particular has a structural blind spot.
\begin{remark}\label{rem:average}
A linear readout of the time-averaged latent, $w^\top \bar\vz$ with $\bar\vz = \frac1T \sum_t \vz_t$, depends on a video only through $\bar\vz$. It cannot separate videos whose frames differ by a rearrangement in time, or by any change that leaves the average unchanged.
\end{remark}
Impossible events such as a vanishing or teleporting object are mostly changes over time, which an average over all frames largely removes, so chance accuracy of this baseline is no evidence that the latent lacks the signal. The remark does not explain why linear readouts that keep the time axis fail too. The magnitude of change, which they cannot form, is what separates the videos here. Plausibility is therefore not linearly decodable from the clean latent, in agreement with their baseline, but it is present there, and a fixed, training-free transform exposes it. Kinematic statistics of frame-to-frame change likewise detect physics violations in the features of frozen image encoders, where a linear probe on the same features reaches 55.5\% on IntPhys~2 \citep{interno2026geophys}; here the same idea applies to a generator's own VAE latent. By the argument of Section~\ref{sec:fair}, no transformer layer can carry more information about it than this latent together with the text prompt. A transformer that makes the signal linearly decodable has made it more usable, as a good representation should. That alone does not show that the model constructs knowledge absent from its input. Our test does not show that the transformer adds nothing: its per-video accuracy may well exceed our 62\% (they report 81.27\% per-video for the 1.3B Wan model, averaged over transformer blocks and over the categories of IntPhys and a second benchmark, InfLevel).

\section{Discussion}
\label{sec:discussion}

\paragraph{What ``knows more than it shows'' can mean.} In the information sense the claim is always true. A diffusion model computes its clean-sample estimate from its activations and its other inputs, so these hold at least as much as the estimate shows. The claim worth testing is narrower: that the activations hold usable information about the final sample that a readout of the model's own estimate, equally flexible and equally well tuned, cannot extract. We expected to find such information and found almost none. The layers did look ahead under the standard comparison, and the advantage went away once the estimate was read with the same care, apart from a small gain for floating in the smaller main data sets. What remains is the fall rate of jumps. The layers carry it early in denoising and the estimate shows it only later; on the 50-step model the gain survives every control we applied, and on every model it is small and short-lived.

In practical terms, the floating motions and the balance margins of MDM's samples are as visible in its own estimate after a fifth of the denoising steps as in its layers. Corrections and filters can act on that estimate directly. How fast a jump falls is the one place where the layers have more to offer, and only early in denoising.

\paragraph{Recommendations for probing studies.} Our results suggest five checks before concluding that a generator knows more than it shows:
\begin{enumerate}
    \item Read the model's own clean-sample estimate at the same step as carefully as the activations: the same readout family, tuned the same way. Report what the layer adds to the estimate, for every layer and with an estimator that can fall back to the estimate alone, and run a positive control on the score the conclusion rests on. Near its ceiling, a pooled AUC can miss a real gain.
    \item Score within the groups that share a prompt or a scene, and within prompt and frame where timing is shared, so that recognizing the condition cannot carry the result. Report how the comparison changes with the amount of training data: with a few hundred examples, a high-dimensional probe and a one-number baseline can trade places.
    \item Before concluding that a property is absent from the input, try simple fixed transforms of it as well as linear probes. In a video latent where every linear readout is at chance, the magnitude of local frame-to-frame change exposes plausibility.
    \item Look at what a target's variance is made of. A gravity score can mostly measure whether a body floats or falls, which the estimate shows early, instead of how fast it falls, which it shows only later.
    \item For early pruning, compare with scoring the decoded preview, and report the saving a perfect detector would give.
\end{enumerate}

\paragraph{Limitations.} We study kinematic generators: two transformer diffusion models trained on HumanML3D, MDM at two checkpoints and MotionDiffuse. Models with other training data or designs, such as Kimodo \citep{rempe2026kimodo} or tokenized generators, may behave differently. Physics-based generators are outside the question, since gravity and contact hold in them by construction and our labels cannot fail. What the kinematic planner of a hybrid represents is a separate question (below). Labels are computed from joints with heuristic thresholds (contact heights, the 15\,cm support rule, the 2\,m/s$^2$ floating cut-off). On real motion capture they flag almost no floating and recover the fall rate of real jumps with a spread of about 0.7 to 1.0\,m/s$^2$ (Appendix~\ref{app:realmotion}), but we checked one dataset of scripted actions, and the balance margins are probed as continuous quantities rather than failure labels. Our conclusion that the layers add nothing is relative to the readout families we tried, ridge and tuned MLPs on one frame at a time and a small temporal network. A stronger readout could change the comparison. The data-processing argument does not protect this conclusion: it rules out that the estimate holds information the activations and noisy input lack, but allows the reverse, so it is the layer side that a stronger readout could still raise. The fall-rate result comes from jump-heavy runs, and we do not know which computation in the layers carries it, or whether steering along it would change how fast the generated jumps fall. On MDM our probes see only the conditional pass of classifier-free guidance. The early-warning study is simulated from recorded samples. Each decision uses exactly what an online run would see, but retries draw from a finite pool of 12 recorded samples per prompt. The video test reads only the clean VAE latent of the IntPhys development set. It tests the evidence for ``absent from the input'', not the transformer's own advantage, and Esmati et al.'s latent baseline is reconstructed from their description because their public code does not include it. We do not test whether the model causally uses the directions we decode.

\paragraph{Future work.} The natural next test is an on-trajectory replication in a video diffusion model, with a physics label measurable from the generated video (for example the fall rate of a tracked object, a quantity that \citet{kneifl2026physics} probe with simulator labels) and the model's decoded preview as the baseline. The same comparison applies to the kinematic planners inside physics-based hybrids \citep{tevet2024closd} and to models that drive humanoid robots \citep{wang2026motionbricks}, where knowing whether the planner's own estimate already reveals an imbalance would tell the downstream controller when to intervene.

\bibliography{references}

\begin{thebibliography}{52}
\providecommand{\natexlab}[1]{#1}
\providecommand{\url}[1]{\texttt{#1}}
\expandafter\ifx\csname urlstyle\endcsname\relax
  \providecommand{\doi}[1]{doi: #1}\else
  \providecommand{\doi}{doi: \begingroup \urlstyle{rm}\Url}\fi

\bibitem[Alain \& Bengio(2016)Alain and Bengio]{alain2016understanding}
Guillaume Alain and Yoshua Bengio.
\newblock Understanding intermediate layers using linear classifier probes.
\newblock \emph{arXiv preprint arXiv:1610.01644}, 2016.

\bibitem[Alam(2026)]{alam2026steering}
Nahid Alam.
\newblock Causal physics steering in video world models via concept activation
  vectors.
\newblock \emph{arXiv preprint arXiv:2605.24322}, 2026.

\bibitem[Belinkov(2022)]{belinkov2022probing}
Yonatan Belinkov.
\newblock Probing classifiers: Promises, shortcomings, and advances.
\newblock \emph{Computational Linguistics}, 48\penalty0 (1), 2022.
\newblock arXiv:2102.12452.

\bibitem[Bogo et~al.(2016)Bogo, Kanazawa, Lassner, Gehler, Romero, and
  Black]{bogo2016keep}
Federica Bogo, Angjoo Kanazawa, Christoph Lassner, Peter Gehler, Javier Romero,
  and Michael~J. Black.
\newblock Keep it {SMPL}: Automatic estimation of {3D} human pose and shape
  from a single image.
\newblock In \emph{European Conference on Computer Vision (ECCV)}, 2016.
\newblock arXiv:1607.08128.

\bibitem[Chen et~al.(2023{\natexlab{a}})Chen, Jiang, Liu, Huang, Fu, Chen, Yu,
  and Yu]{chen2023mld}
Xin Chen, Biao Jiang, Wen Liu, Zilong Huang, Bin Fu, Tao Chen, Jingyi Yu, and
  Gang Yu.
\newblock Executing your commands via motion diffusion in latent space.
\newblock In \emph{Proceedings of the IEEE/CVF Conference on Computer Vision
  and Pattern Recognition (CVPR)}, 2023{\natexlab{a}}.
\newblock arXiv:2212.04048.

\bibitem[Chen et~al.(2023{\natexlab{b}})Chen, Vi{\'e}gas, and
  Wattenberg]{chen2023beyond}
Yida Chen, Fernanda Vi{\'e}gas, and Martin Wattenberg.
\newblock Beyond surface statistics: Scene representations in a latent
  diffusion model.
\newblock \emph{arXiv preprint arXiv:2306.05720}, 2023{\natexlab{b}}.

\bibitem[Cui et~al.(2026)Cui, Huang, Ye, Dong, Chen, Xue, Yang, Huang, Tang,
  and Hong]{cui2026diffusionprobe}
Benlei Cui, Bukun Huang, Zhizeng Ye, Xuemei Dong, Tuo Chen, Hui Xue, Dingkang
  Yang, Longtao Huang, Jingqun Tang, and Haiwen Hong.
\newblock Diffusion probe: Generated image result prediction using {CNN}
  probes.
\newblock \emph{arXiv preprint arXiv:2602.23783}, 2026.

\bibitem[Elazar et~al.(2021)Elazar, Ravfogel, Jacovi, and
  Goldberg]{elazar2021amnesic}
Yanai Elazar, Shauli Ravfogel, Alon Jacovi, and Yoav Goldberg.
\newblock Amnesic probing: Behavioral explanation with amnesic counterfactuals.
\newblock \emph{Transactions of the Association for Computational Linguistics},
  2021.
\newblock arXiv:2006.00995.

\bibitem[Esmati et~al.(2026)Esmati, Nath, Hofmann, Nowrouzezahrai, Kahou, and
  Mirmehdi]{esmati2026invisible}
Parsa Esmati, Somjit Nath, Katja Hofmann, Derek Nowrouzezahrai, Samira~Ebrahimi
  Kahou, and Majid Mirmehdi.
\newblock The invisible hand of physics: When video diffusion models know more
  than they show.
\newblock \emph{arXiv preprint arXiv:2606.05328}, 2026.

\bibitem[Garrido et~al.(2025)Garrido, Ballas, Assran, Bardes, Najman, Rabbat,
  Dupoux, and LeCun]{garrido2025intuitive}
Quentin Garrido, Nicolas Ballas, Mahmoud Assran, Adrien Bardes, Laurent Najman,
  Michael Rabbat, Emmanuel Dupoux, and Yann LeCun.
\newblock Intuitive physics understanding emerges from self-supervised
  pretraining on natural videos.
\newblock \emph{arXiv preprint arXiv:2502.11831}, 2025.

\bibitem[Ghorbani et~al.(2021)Ghorbani, Mahdaviani, Thaler, Kording, Cook,
  Blohm, and Troje]{ghorbani2021movi}
Saeed Ghorbani, Kimia Mahdaviani, Anne Thaler, Konrad Kording, Douglas~James
  Cook, Gunnar Blohm, and Nikolaus~F. Troje.
\newblock {MoVi}: A large multi-purpose human motion and video dataset.
\newblock \emph{PLOS ONE}, 16\penalty0 (6):\penalty0 e0253157, 2021.

\bibitem[Guo et~al.(2022)Guo, Zou, Zuo, Wang, Ji, Li, and
  Cheng]{guo2022humanml3d}
Chuan Guo, Shihao Zou, Xinxin Zuo, Sen Wang, Wei Ji, Xingyu Li, and Li~Cheng.
\newblock Generating diverse and natural {3D} human motions from text.
\newblock In \emph{Proceedings of the IEEE/CVF Conference on Computer Vision
  and Pattern Recognition (CVPR)}, 2022.

\bibitem[Guo et~al.(2024)Guo, Mu, Javed, Wang, and Cheng]{guo2024momask}
Chuan Guo, Yuxuan Mu, Muhammad~Gohar Javed, Sen Wang, and Li~Cheng.
\newblock {MoMask}: Generative masked modeling of {3D} human motions.
\newblock In \emph{Proceedings of the IEEE/CVF Conference on Computer Vision
  and Pattern Recognition (CVPR)}, 2024.

\bibitem[Guo et~al.(2026)Guo, Wei, and Jing]{guo2026early}
Huanlei Guo, Hongxin Wei, and Bingyi Jing.
\newblock Toward early quality assessment of text-to-image diffusion models.
\newblock \emph{arXiv preprint arXiv:2603.02829}, 2026.

\bibitem[Hewitt \& Liang(2019)Hewitt and Liang]{hewitt2019designing}
John Hewitt and Percy Liang.
\newblock Designing and interpreting probes with control tasks.
\newblock In \emph{Proceedings of the 2019 Conference on Empirical Methods in
  Natural Language Processing and the 9th International Joint Conference on
  Natural Language Processing (EMNLP-IJCNLP)}, 2019.

\bibitem[Hewitt et~al.(2021)Hewitt, Ethayarajh, Liang, and
  Manning]{hewitt2021conditional}
John Hewitt, Kawin Ethayarajh, Percy Liang, and Christopher~D. Manning.
\newblock Conditional probing: Measuring usable information beyond a baseline.
\newblock In \emph{Proceedings of the 2021 Conference on Empirical Methods in
  Natural Language Processing (EMNLP)}, 2021.
\newblock arXiv:2109.09234.

\bibitem[Hof et~al.(2005)Hof, Gazendam, and Sinke]{hof2005condition}
A.~L. Hof, M.~G.~J. Gazendam, and W.~E. Sinke.
\newblock The condition for dynamic stability.
\newblock \emph{Journal of Biomechanics}, 38\penalty0 (1):\penalty0 1--8, 2005.

\bibitem[Holden et~al.(2016)Holden, Saito, and Komura]{holden2016deep}
Daniel Holden, Jun Saito, and Taku Komura.
\newblock A deep learning framework for character motion synthesis and editing.
\newblock \emph{ACM Transactions on Graphics}, 35\penalty0 (4):\penalty0
  138:1--138:11, 2016.
\newblock \doi{10.1145/2897824.2925975}.

\bibitem[Holden et~al.(2017)Holden, Komura, and Saito]{holden2017pfnn}
Daniel Holden, Taku Komura, and Jun Saito.
\newblock Phase-functioned neural networks for character control.
\newblock \emph{ACM Transactions on Graphics}, 36\penalty0 (4):\penalty0
  42:1--42:13, 2017.
\newblock \doi{10.1145/3072959.3073663}.

\bibitem[Intern{\`o} et~al.(2026)Intern{\`o}, Pondaven, Issa, Pizzati, Pinto,
  Olhofer, Laptev, Torr, Simoncelli, Hammer, and Klindt]{interno2026geophys}
Christian Intern{\`o}, Alexander Pondaven, Habon Issa, Fabio Pizzati, Francesco
  Pinto, Markus Olhofer, Ivan Laptev, Philip Torr, Eero~P. Simoncelli, Barbara
  Hammer, and David Klindt.
\newblock {GEOPHYS}: The geometry of physical plausibility.
\newblock \emph{arXiv preprint arXiv:2606.20707}, 2026.

\bibitem[Joseph et~al.(2026)Joseph, Garrido, Balestriero, Kowal, Fel,
  Bakhtiari, Richards, and Rabbat]{joseph2026interpreting}
Sonia Joseph, Quentin Garrido, Randall Balestriero, Matthew Kowal, Thomas Fel,
  Shahab Bakhtiari, Blake Richards, and Mike Rabbat.
\newblock Interpreting physics in video world models.
\newblock In \emph{Proceedings of the 43rd International Conference on Machine
  Learning (ICML)}, 2026.
\newblock arXiv:2602.07050.

\bibitem[Karunratanakul et~al.(2023)Karunratanakul, Preechakul, Suwajanakorn,
  and Tang]{karunratanakul2023gmd}
Korrawe Karunratanakul, Konpat Preechakul, Supasorn Suwajanakorn, and Siyu
  Tang.
\newblock Guided motion diffusion for controllable human motion synthesis.
\newblock In \emph{Proceedings of the IEEE/CVF International Conference on
  Computer Vision (ICCV)}, 2023.
\newblock arXiv:2305.12577.

\bibitem[Karunratanakul et~al.(2024)Karunratanakul, Preechakul, Aksan, Beeler,
  Suwajanakorn, and Tang]{karunratanakul2024dno}
Korrawe Karunratanakul, Konpat Preechakul, Emre Aksan, Thabo Beeler, Supasorn
  Suwajanakorn, and Siyu Tang.
\newblock Optimizing diffusion noise can serve as universal motion priors.
\newblock In \emph{Proceedings of the IEEE/CVF Conference on Computer Vision
  and Pattern Recognition (CVPR)}, 2024.
\newblock arXiv:2312.11994.

\bibitem[Kneifl et~al.(2026)Kneifl, Skalski, Twardowski, and
  Deja]{kneifl2026physics}
Jonas Kneifl, Jakub Skalski, Bart{\l}omiej Twardowski, and Kamil Deja.
\newblock Does physics live in the activations? {L}ocalizing physical
  quantities in video diffusion models.
\newblock \emph{arXiv preprint arXiv:2610.03154}, 2026.

\bibitem[Loper et~al.(2015)Loper, Mahmood, Romero, Pons-Moll, and
  Black]{loper2015smpl}
Matthew Loper, Naureen Mahmood, Javier Romero, Gerard Pons-Moll, and Michael~J.
  Black.
\newblock {SMPL}: A skinned multi-person linear model.
\newblock \emph{ACM Transactions on Graphics (Proc. SIGGRAPH Asia)},
  34\penalty0 (6):\penalty0 248:1--248:16, 2015.

\bibitem[Mahmood et~al.(2019)Mahmood, Ghorbani, Troje, Pons-Moll, and
  Black]{mahmood2019amass}
Naureen Mahmood, Nima Ghorbani, Nikolaus~F. Troje, Gerard Pons-Moll, and
  Michael~J. Black.
\newblock {AMASS}: Archive of motion capture as surface shapes.
\newblock In \emph{Proceedings of the IEEE/CVF International Conference on
  Computer Vision (ICCV)}, 2019.
\newblock arXiv:1904.03278.

\bibitem[Musa et~al.(2026)Musa, Forghani, Thasarathan, Joseph, Kowal, and
  Derpanis]{musa2026whatwhere}
Sharon~S. Musa, Fereshteh Forghani, Harrish Thasarathan, Sonia Joseph, Matthew
  Kowal, and Konstantinos~G. Derpanis.
\newblock What, where, and how: Probing spatiotemporal representations in video
  foundation models.
\newblock \emph{arXiv preprint arXiv:2609.01551}, 2026.

\bibitem[Peng et~al.(2018)Peng, Abbeel, Levine, and van~de
  Panne]{peng2018deepmimic}
Xue~Bin Peng, Pieter Abbeel, Sergey Levine, and Michiel van~de Panne.
\newblock {DeepMimic}: Example-guided deep reinforcement learning of
  physics-based character skills.
\newblock \emph{ACM Transactions on Graphics}, 37\penalty0 (4), 2018.

\bibitem[Pimentel et~al.(2020{\natexlab{a}})Pimentel, Saphra, Williams, and
  Cotterell]{pimentel2020pareto}
Tiago Pimentel, Naomi Saphra, Adina Williams, and Ryan Cotterell.
\newblock Pareto probing: Trading off accuracy for complexity.
\newblock In \emph{Proceedings of the 2020 Conference on Empirical Methods in
  Natural Language Processing (EMNLP)}, 2020{\natexlab{a}}.
\newblock arXiv:2010.02180.

\bibitem[Pimentel et~al.(2020{\natexlab{b}})Pimentel, Valvoda, Hall~Maudslay,
  Zmigrod, Williams, and Cotterell]{pimentel2020information}
Tiago Pimentel, Josef Valvoda, Rowan Hall~Maudslay, Ran Zmigrod, Adina
  Williams, and Ryan Cotterell.
\newblock Information-theoretic probing for linguistic structure.
\newblock In \emph{Proceedings of the 58th Annual Meeting of the Association
  for Computational Linguistics (ACL)}, 2020{\natexlab{b}}.
\newblock arXiv:2004.03061.

\bibitem[Punzo et~al.(2026)Punzo, Caselli, Pantelidis, Massafra, Lo~Sardo, and
  Salehi]{punzo2026video}
Samuele Punzo, Niccol{\`o} Caselli, Ippokratis Pantelidis, Francesco Massafra,
  Salvatore Lo~Sardo, and Mohammadreza Salehi.
\newblock How do video foundation models encode intuitive physics? {P}robing
  across pretraining paradigms.
\newblock \emph{arXiv preprint arXiv:2606.09646}, 2026.

\bibitem[Rempe et~al.(2026)Rempe, Petrovich, Yuan, Zhang, Peng, Jiang, Wang,
  Iqbal, Minor, de~Ruyter, Li, Tessler, Lim, Jeong, Wu, Hassani, Huang, Yu,
  Chung, Song, Dionne, Kautz, Yuen, and Fidler]{rempe2026kimodo}
Davis Rempe, Mathis Petrovich, Ye~Yuan, Haotian Zhang, Xue~Bin Peng, Yifeng
  Jiang, Tingwu Wang, Umar Iqbal, David Minor, Michael de~Ruyter, Jiefeng Li,
  Chen Tessler, Edy Lim, Eugene Jeong, Sam Wu, Ehsan Hassani, Michael Huang,
  Jin-Bey Yu, Chaeyeon Chung, Lina Song, Olivier Dionne, Jan Kautz, Simon Yuen,
  and Sanja Fidler.
\newblock Kimodo: Scaling controllable human motion generation.
\newblock \emph{arXiv preprint arXiv:2603.15546}, 2026.

\bibitem[Riochet et~al.(2021)Riochet, Ynocente~Castro, Bernard, Lerer, Fergus,
  Izard, and Dupoux]{riochet2018intphys}
Ronan Riochet, Mario Ynocente~Castro, Mathieu Bernard, Adam Lerer, Rob Fergus,
  V{\'e}ronique Izard, and Emmanuel Dupoux.
\newblock {IntPhys} 2019: A benchmark for visual intuitive physics
  understanding.
\newblock \emph{IEEE Transactions on Pattern Analysis and Machine
  Intelligence}, 2021.
\newblock \doi{10.1109/TPAMI.2021.3083839}.

\bibitem[Romero et~al.(2017)Romero, Tzionas, and Black]{romero2017mano}
Javier Romero, Dimitrios Tzionas, and Michael~J. Black.
\newblock Embodied hands: Modeling and capturing hands and bodies together.
\newblock \emph{ACM Transactions on Graphics}, 36\penalty0 (6):\penalty0 245,
  2017.

\bibitem[Starke et~al.(2019)Starke, Zhang, Komura, and Saito]{starke2019nsm}
Sebastian Starke, He~Zhang, Taku Komura, and Jun Saito.
\newblock Neural state machine for character-scene interactions.
\newblock \emph{ACM Transactions on Graphics}, 38\penalty0 (6):\penalty0
  178:1--178:14, 2019.
\newblock \doi{10.1145/3355089.3356505}.

\bibitem[St{\"o}lzle et~al.(2026)St{\"o}lzle, M{\"u}ller, Serifi, Grandia,
  Knoop, and B{\"a}cher]{physiblemdm2026}
Maximilian St{\"o}lzle, David M{\"u}ller, Agon Serifi, Ruben Grandia, Espen
  Knoop, and Moritz B{\"a}cher.
\newblock {PhysibleMDM}: Enforcing inference-time physical feasibility
  constraints on motion diffusion models via noise optimization.
\newblock \emph{arXiv preprint arXiv:2610.30500}, 2026.

\bibitem[Tang et~al.(2026)Tang, Zhong, and Ding]{tang2026seeking}
Chujun Tang, Lei Zhong, and Fangqiang Ding.
\newblock Seeking physics in diffusion noise.
\newblock \emph{arXiv preprint arXiv:2603.14294}, 2026.

\bibitem[{Team Wan} et~al.(2025){Team Wan}, Wang, Ai, et~al.]{wan2025}
{Team Wan}, Ang Wang, Baole Ai, et~al.
\newblock Wan: Open and advanced large-scale video generative models.
\newblock \emph{arXiv preprint arXiv:2503.20314}, 2025.

\bibitem[Tessler et~al.(2024)Tessler, Guo, Nabati, Chechik, and
  Peng]{tessler2024maskedmimic}
Chen Tessler, Yunrong Guo, Ofir Nabati, Gal Chechik, and Xue~Bin Peng.
\newblock {MaskedMimic}: Unified physics-based character control through masked
  motion inpainting.
\newblock \emph{ACM Transactions on Graphics}, 43\penalty0 (6):\penalty0 1--21,
  2024.

\bibitem[Tevet et~al.(2023)Tevet, Raab, Gordon, Shafir, Cohen-Or, and
  Bermano]{tevet2023mdm}
Guy Tevet, Sigal Raab, Brian Gordon, Yoni Shafir, Daniel Cohen-Or, and
  Amit~Haim Bermano.
\newblock Human motion diffusion model.
\newblock In \emph{International Conference on Learning Representations
  (ICLR)}, 2023.

\bibitem[Tevet et~al.(2024)Tevet, Raab, Cohan, Reda, Luo, Peng, Bermano, and
  van~de Panne]{tevet2024closd}
Guy Tevet, Sigal Raab, Setareh Cohan, Daniele Reda, Zhengyi Luo, Xue~Bin Peng,
  Amit~H. Bermano, and Michiel van~de Panne.
\newblock {CLoSD}: Closing the loop between simulation and diffusion for
  multi-task character control.
\newblock \emph{arXiv preprint arXiv:2410.03441}, 2024.

\bibitem[Voita \& Titov(2020)Voita and Titov]{voita2020mdl}
Elena Voita and Ivan Titov.
\newblock Information-theoretic probing with minimum description length.
\newblock In \emph{Proceedings of the 2020 Conference on Empirical Methods in
  Natural Language Processing (EMNLP)}, 2020.
\newblock arXiv:2003.12298.

\bibitem[Wang et~al.(2026)Wang, Dionne, de~Ruyter, Minor, Rempe, Zhao,
  Petrovich, Yuan, Li, Luo, Robison, Blackwell, Antoniazzi, Peng, Zhu, and
  Yuen]{wang2026motionbricks}
Tingwu Wang, Olivier Dionne, Michael de~Ruyter, David Minor, Davis Rempe,
  Kaifeng Zhao, Mathis Petrovich, Ye~Yuan, Chenran Li, Zhengyi Luo, Brian
  Robison, Xavier Blackwell, Bernardo Antoniazzi, Xue~Bin Peng, Yuke Zhu, and
  Simon Yuen.
\newblock {MotionBricks}: Scalable real-time motions with modular latent
  generative model and smart primitives.
\newblock \emph{ACM Transactions on Graphics}, 45\penalty0 (4), 2026.
\newblock \doi{10.1145/3811334}.

\bibitem[Watanabe et~al.(2025)Watanabe, Ren, Siyao, Peng, Wu, and
  Simo-Serra]{simdiff2025}
Akihisa Watanabe, Jiawei Ren, Li~Siyao, Yichen Peng, Erwin Wu, and Edgar
  Simo-Serra.
\newblock {SimDiff}: Simulator-constrained diffusion model for physically
  plausible motion generation.
\newblock \emph{arXiv preprint arXiv:2509.20927}, 2025.

\bibitem[Winter(2009)]{winter2009biomechanics}
David~A. Winter.
\newblock \emph{Biomechanics and Motor Control of Human Movement}.
\newblock Wiley, 4th edition, 2009.

\bibitem[Xie et~al.(2024)Xie, Jampani, Zhong, Sun, and
  Jiang]{xie2024omnicontrol}
Yiming Xie, Varun Jampani, Lei Zhong, Deqing Sun, and Huaizu Jiang.
\newblock {OmniControl}: Control any joint at any time for human motion
  generation.
\newblock In \emph{International Conference on Learning Representations
  (ICLR)}, 2024.
\newblock arXiv:2310.08580.

\bibitem[Xing et~al.(2026)Xing, Mao, and Liu]{xing2026interphys}
Chaoyue Xing, Wei Mao, and Miaomiao Liu.
\newblock {InterPhys}: Physics-aware human motion synthesis in a dynamic scene.
\newblock In \emph{Proceedings of the IEEE/CVF Conference on Computer Vision
  and Pattern Recognition (CVPR)}, pp.\  30729--30739, 2026.

\bibitem[Xu et~al.(2020)Xu, Zhao, Song, Stewart, and Ermon]{xu2020vinfo}
Yilun Xu, Shengjia Zhao, Jiaming Song, Russell Stewart, and Stefano Ermon.
\newblock A theory of usable information under computational constraints.
\newblock In \emph{International Conference on Learning Representations
  (ICLR)}, 2020.
\newblock arXiv:2002.10689.

\bibitem[Yuan et~al.(2023)Yuan, Song, Iqbal, Vahdat, and
  Kautz]{yuan2023physdiff}
Ye~Yuan, Jiaming Song, Umar Iqbal, Arash Vahdat, and Jan Kautz.
\newblock {PhysDiff}: Physics-guided human motion diffusion model.
\newblock In \emph{Proceedings of the IEEE/CVF International Conference on
  Computer Vision (ICCV)}, 2023.

\bibitem[Zhang et~al.(2023)Zhang, Zhang, Cun, Huang, Zhang, Zhao, Lu, and
  Shen]{zhang2023t2mgpt}
Jianrong Zhang, Yangsong Zhang, Xiaodong Cun, Shaoli Huang, Yong Zhang, Hongwei
  Zhao, Hongtao Lu, and Xi~Shen.
\newblock {T2M-GPT}: Generating human motion from textual descriptions with
  discrete representations.
\newblock In \emph{Proceedings of the IEEE/CVF Conference on Computer Vision
  and Pattern Recognition (CVPR)}, 2023.

\bibitem[Zhang et~al.(2022)Zhang, Cai, Pan, Hong, Guo, Yang, and
  Liu]{zhang2022motiondiffuse}
Mingyuan Zhang, Zhongang Cai, Liang Pan, Fangzhou Hong, Xinying Guo, Lei Yang,
  and Ziwei Liu.
\newblock {MotionDiffuse}: Text-driven human motion generation with diffusion
  model.
\newblock \emph{arXiv preprint arXiv:2208.15001}, 2022.

\bibitem[Zhang et~al.(2026)Zhang, Leng, Bao, Zhao, Zhou, and
  Zhou]{zhang2026readability}
Xiang Zhang, Ziyang Leng, Yuxin Bao, Zhihao Zhao, Bolei Zhou, and Xugui Zhou.
\newblock Early readability of trajectory quality in diffusion driving
  planners.
\newblock \emph{arXiv preprint arXiv:2611.19684}, 2026.

\end{thebibliography}
\bibliographystyle{tmlr}

\appendix
\section{Implementation details}
\label{app:impl}

\paragraph{Model and checkpoints.} We use the official MDM code \citep{tevet2023mdm} and its HumanML3D encoder checkpoints with 50 and 1000 diffusion steps. Both have the same architecture and text encoder, but they were trained with different releases of the code, and their saved training settings differ beyond the number of steps (each checkpoint's \texttt{args.json}); we therefore treat them as two models, not as a controlled comparison of step counts. Joint positions are recovered from the 263-dimensional HumanML3D features with the dataset's normalization statistics; no SMPL body model is needed. Each layer's output is recorded with a forward hook after the condition token is dropped; in the main runs we keep every fourth frame (30 per motion), and the noisy input and the clean-motion estimate are stored at the same frames.

\paragraph{Compute.} All experiments ran on a single laptop GPU (Apple M4 Pro). A forward pass on the laptop GPU matches the CPU to about $10^{-4}$ relative error in every layer. Sampling 1,000 motions takes about 2 minutes with the 50-step model and under an hour with the 1000-step model. The SMPL bodies in the figures are fitted to the generated joints for display only, with mean joint errors of 1.0 to 3.0\,cm per figure (largest for two of the extreme jumps in Figure~\ref{fig:layerknows}c). All probes, controls and simulations run on cached arrays in minutes to an hour. Code that reproduces every experiment, table and figure, from sampling to the final plots, is available at \url{https://github.com/saeed1262/motion-physics-probing}.

\paragraph{Body meshes in Figure~\ref{fig:teaser}.} For display only, we fit the neutral SMPL body \citep{loper2015smpl} to the 22 joints of each drawn frame with the SMPLify-style fitter distributed with MDM \citep{bogo2016keep} (150 iterations). The mean distance between fitted and generated joints is 1.0 to 1.2\,cm across the six rows. The lowest mesh vertex lies 1.2 to 6.5\,cm below the lowest foot joint, as expected since the ankle and toe joints sit inside the foot, so a floating body's sole stays visibly above its shadow.

\paragraph{Prompts.} The 64 curated prompts (each preceded by ``a person'') are:
{\small walks forward; walks forward slowly and carefully; jogs forward; runs forward quickly; walks in a circle; walks backward; sidesteps to the left; walks forward and turns around; jumps up in place; jumps forward; does a long jump forward; hops on their left leg; hops on their right leg; jumps over an obstacle; skips forward; does jumping jacks; jumps up and spins in the air; leaps to the side; runs and jumps; jumps backward; jumps with both feet three times; jumps and lands on one foot; stands on one leg; balances on their right foot with arms out; leans forward; leans to the left; stands on their toes; walks on a balance beam; loses balance and catches themselves; stumbles forward; bends down to touch their toes; squats down and stands up; does a lunge forward; kicks with their right leg; kicks high with their left leg; does a yoga tree pose; tiptoes forward; runs and then stops suddenly; waves with their right hand; throws a ball; picks something up from the ground; reaches up high; claps their hands; stretches their arms; punches forward; boxes; swings a golf club; kicks a soccer ball; dances; does a spin; marches in place; crouches and walks; walks while carrying something heavy; dribbles a basketball and shoots; plays tennis; shuffles side to side; steps forward and back; turns in place; stands still; stretches their legs; shakes their body; runs in a zigzag; jogs in place; sprints.}

The 186 test captions are drawn at random (fixed seed) from the HumanML3D test split, one caption per motion, keeping captions of 3 to 20 words that do not duplicate a curated prompt.

\paragraph{Readouts.} Per-frame probes predict the final motion's label at the same frame. Ridge probes use standardized inputs and choose the penalty from $\{1, 10, 100, 1000, 10^4\}$ by efficient leave-one-out cross-validation. The \emph{Estimate, MLP} readout concatenates the estimate's 263 features with the dynamic and static balance margins, lowest-foot height, in-flight and airborne flags and foot sliding speed computed from the estimate (and, for the gravity target, its flight gravity), each with an indicator for undefined values (undefined values set to zero), and its four foot-contact flags. Where the estimate's own label is undefined, the explicit readout uses the median of its defined values. The MLP (two hidden layers of 256 units, ReLU, dropout 0.1) is trained on standardized inputs and targets with AdamW (learning rate $10^{-3}$, weight decay $10^{-4}$, batch 512) for up to 100 epochs, stopping after 8 epochs without improvement on 15\% of the training prompts, held out as a group; the same settings are used for every feature set. Fixed layers: dynamic and static balance, layer 7; flight, layer 6; flight gravity, layer 5.

\paragraph{Temporal readout.} To test whether per-frame readouts handicap the estimate on quantities that need several frames, a small temporal network reads all 30 recorded frames of a motion and predicts the final motion's label at every frame: two 1D convolutions over time (128 channels, kernel 5, the second dilated by 2, GELU, dropout 0.1) and a per-frame linear output, trained with AdamW (learning rate $10^{-3}$, weight decay $10^{-3}$, batch 32) on a loss masked to labelled frames, with early stopping (patience 15, up to 200 epochs) on 15\% of the training prompts. Inputs are everything the estimate shows, the same fixed layer, or both, at 20\% of denoising; folds and scores are those of the per-frame readouts, one seed at a time. Trained per seed, it reads dynamic balance better from the estimate than from the layer in all six runs (by 0.014 to 0.034 in $R^2$), and flight as well (by 0.001 to 0.010 in AUC). For the fall rate on the jump prompts it reaches $R^2 = 0.297 \pm 0.017$ from the estimate and $0.379 \pm 0.025$ from the layer on the 50-step model, and $0.290 \pm 0.114$ and $0.344 \pm 0.124$ on the 1000-step model, with the layer ahead in every seed (by 0.038 to 0.097).

\paragraph{Tuned readouts.} With seeds pooled, the five folds are grouped by prompt across seeds, and inside each outer fold 15\% of the training prompts form the validation split used for every choice. Ridge selects its penalty from $\{10^{-1}, 1, \dots, 10^{6}\}$ on that split (inputs standardized, targets centred) and is refit on all training frames. The MLP family is $f(x) = w^\top x + b + \mathrm{MLP}(x)$, so that it contains the linear readout, trained on standardized inputs and targets with AdamW (learning rate $10^{-3}$, batch 1024) for up to 80 epochs, stopping after 6 epochs without improvement on the validation split. Its five configurations are (hidden units, hidden layers, dropout, weight decay): $(256, 2, 0.1, 10^{-4})$, the fixed MLP's settings; $(256, 2, 0.3, 10^{-2})$; $(64, 2, 0.1, 10^{-3})$; $(1024, 1, 0.3, 10^{-2})$; and $(1024, 2, 0.3, 10^{-3})$. The configuration and the family with the lowest validation error are kept, separately in each outer fold. Ridge is then refit on all training frames, while the MLP keeps the fit on the 85\% used with early stopping. Selection uses mean squared error for every target, including flight and floats or falls, which are scored by AUC. The MLP family was selected in 70 to 100\% of folds per data set, and every configuration at least once. The same fixed layers are used as above for MDM, and Section~\ref{sec:models} gives MotionDiffuse's. The readout of estimate and layer with fallback chooses among the 12 candidates of both input sets (two ridge fits and ten MLPs over the two inputs, counting the estimate's) by the same validation error. Over MDM's four data sets it picks the estimate alone in 25 of the 40 folds for balance and flight, and in 1 of the 10 for the fall rate on the jump prompts. Our per-frame inputs omit the condition token (token 0 of the encoder). To test it, we recorded the 50-step model's three main runs again with the token kept (the motions, and the estimate's scores, are identical); adding the token's output at the same layer (the same for every frame of a motion) to the readout of estimate and layer gains $-0.001$ [$-0.003$, $0.002$] for dynamic balance, $+0.004$ [$-0.000$, $0.009$] for flight and $+0.002$ [$-0.003$, $0.007$] for whether a body floats or falls (tuned with fallback; with all eight layers, $0.000$, $0.000$ and $-0.007$). Table~\ref{tab:raw} separates the two ways in which the estimate's readout differs from the layer's: the input (the estimate's 263 raw features, or everything it shows) and the family. With raw features the tuned estimate ties the tuned layer for balance and flight on MDM; the labels computed from the estimate, which are fixed functions of it, add about 0.02 in $R^2$ for dynamic balance.

\begin{table}[h]
\centering
\caption{Pooled scores at 20\% of denoising, seeds pooled: the estimate read from its raw features or from everything it shows, and the layer, each with ridge and with the tuned MLP family. $^\dagger$The MLP family diverged in one fold, in every configuration (MotionDiffuse, fall rate, inputs that include the estimate's labels); the tuned readouts chose ridge in that fold, so Table~\ref{tab:tuned} is unaffected.}
\label{tab:raw}
\resizebox{\linewidth}{!}{\begin{tabular}{llcccccc}
\toprule
& & \multicolumn{4}{c}{Estimate} & \multicolumn{2}{c}{Layer} \\
\cmidrule(lr){3-6} \cmidrule(lr){7-8}
Target & Model & raw, ridge & raw, MLP & all it shows, ridge & all it shows, MLP & ridge & MLP \\
\midrule
Dynamic balance & 50-step & 0.767 & 0.869 & 0.866 & 0.889 & 0.850 & 0.862 \\
Dynamic balance & 1000-step & 0.594 & 0.762 & 0.746 & 0.779 & 0.720 & 0.765 \\
Dynamic balance & MotionDiffuse & 0.583 & 0.825 & 0.606 & 0.823 & 0.620 & 0.700 \\
Static balance & 50-step & 0.697 & 0.798 & 0.825 & 0.830 & 0.780 & 0.801 \\
Static balance & 1000-step & 0.544 & 0.646 & 0.641 & 0.671 & 0.592 & 0.647 \\
Static balance & MotionDiffuse & 0.511 & 0.726 & 0.536 & 0.729 & 0.537 & 0.568 \\
In flight & 50-step & 0.944 & 0.975 & 0.974 & 0.976 & 0.977 & 0.971 \\
In flight & 1000-step & 0.903 & 0.933 & 0.919 & 0.935 & 0.934 & 0.932 \\
In flight & MotionDiffuse & 0.963 & 0.944 & 0.967 & 0.943 & 0.951 & 0.898 \\
In flight & 50-step, jump prompts & 0.971 & 0.982 & 0.979 & 0.984 & 0.981 & 0.981 \\
In flight & 1000-step, jump prompts & 0.940 & 0.960 & 0.949 & 0.960 & 0.955 & 0.961 \\
In flight & MotionDiffuse, jump prompts & 0.977 & 0.977 & 0.977 & 0.975 & 0.978 & 0.972 \\
Floats or falls & 50-step & 0.976 & 0.979 & 0.982 & 0.980 & 0.990 & 0.989 \\
Floats or falls & 1000-step & 0.949 & 0.958 & 0.963 & 0.964 & 0.980 & 0.976 \\
Floats or falls & MotionDiffuse & 0.932 & 0.947 & 0.941 & 0.935 & 0.988 & 0.975 \\
Floats or falls & 50-step, jump prompts & 0.952 & 0.962 & 0.978 & 0.974 & 0.971 & 0.969 \\
Floats or falls & 1000-step, jump prompts & 0.958 & 0.963 & 0.962 & 0.966 & 0.973 & 0.969 \\
Floats or falls & MotionDiffuse, jump prompts & 0.833 & 0.781 & 0.841 & 0.826 & 0.836 & 0.788 \\
Fall rate (falling only) & 50-step & $-0.096$ & $-0.372$ & $-0.051$ & 0.031 & 0.267 & 0.217 \\
Fall rate (falling only) & 1000-step & $-0.097$ & $-0.040$ & $-0.055$ & $-0.008$ & 0.115 & 0.054 \\
Fall rate (falling only) & 50-step, jump prompts & 0.183 & 0.267 & 0.273 & 0.351 & 0.374 & 0.427 \\
Fall rate (falling only) & 1000-step, jump prompts & 0.035 & 0.195 & 0.206 & 0.287 & 0.273 & 0.309 \\
Fall rate (falling only) & MotionDiffuse, jump prompts & 0.213 & 0.239 & 0.370 & $-59.7^\dagger$ & 0.326 & 0.231 \\
\bottomrule
\end{tabular}
}
\end{table}

\begin{table}[h]
\centering
\caption{The gain from adding the layer in detail (20\% of denoising, seeds pooled, pooled score, 95\% prompt-bootstrap intervals). With fallback: as in Table~\ref{tab:tuned}. Without fallback: the tuned readout of estimate and layer against the tuned estimate. Estimate kept: outer folds, of five, in which the fallback chose a readout of the estimate alone, so that fold's gain is zero by construction; when all five do, no interval is defined. Rescaled: for $R^2$, the conditional usable information in nats by Proposition~\ref{prop:vinfo}, $\tfrac12 \log\big((1 - R^2_{\text{estimate}})/(1 - R^2_{\text{both}})\big)$; for AUC, the share of misranked pairs that adding the layer removes.}
\label{tab:gaindetail}
\resizebox{\linewidth}{!}{\begin{tabular}{llcccc}
\toprule
Target & Model & Gain with fallback & Gain without fallback & Estimate kept & Gain, rescaled \\
\midrule
Dynamic balance & 50-step & $+0.000$ [$-0.001$, $+0.001$] & $-0.005$ [$-0.010$, $-0.000$] & 4 & +0.001 \\
Dynamic balance & 1000-step & $-0.002$ [$-0.005$, $-0.000$] & $-0.002$ [$-0.006$, $+0.002$] & 3 & $-$0.006 \\
Dynamic balance & MotionDiffuse & $0$ (all folds) & $-0.015$ [$-0.023$, $-0.007$] & 5 & +0.000 \\
\addlinespace[2pt]
Static balance & 50-step & $-0.001$ [$-0.004$, $+0.002$] & $-0.004$ [$-0.008$, $-0.000$] & 2 & $-$0.004 \\
Static balance & 1000-step & $0$ (all folds) & $-0.006$ [$-0.018$, $+0.002$] & 5 & +0.000 \\
Static balance & MotionDiffuse & $0$ (all folds) & $-0.031$ [$-0.051$, $-0.012$] & 5 & +0.000 \\
\addlinespace[2pt]
In flight & 50-step & $-0.004$ [$-0.008$, $-0.000$] & $-0.004$ [$-0.009$, $+0.002$] & 2 & $-$15\% \\
In flight & 1000-step & $-0.000$ [$-0.003$, $+0.002$] & $+0.003$ [$-0.003$, $+0.009$] & 4 & $-$1\% \\
In flight & MotionDiffuse & $-0.004$ [$-0.009$, $-0.001$] & $-0.006$ [$-0.026$, $+0.018$] & 4 & $-$8\% \\
In flight & 50-step, jump prompts & $-0.001$ [$-0.002$, $-0.000$] & $-0.002$ [$-0.004$, $-0.001$] & 4 & $-$5\% \\
In flight & 1000-step, jump prompts & $+0.001$ [$-0.000$, $+0.003$] & $+0.002$ [$+0.000$, $+0.005$] & 1 & +4\% \\
In flight & MotionDiffuse, jump prompts & $+0.000$ [$-0.001$, $+0.002$] & $-0.003$ [$-0.006$, $-0.000$] & 4 & +1\% \\
\addlinespace[2pt]
Floats or falls & 50-step & $+0.008$ [$+0.001$, $+0.015$] & $+0.011$ [$+0.003$, $+0.021$] & 1 & +39\% \\
Floats or falls & 1000-step & $+0.003$ [$-0.001$, $+0.008$] & $+0.011$ [$+0.005$, $+0.019$] & 2 & +10\% \\
Floats or falls & MotionDiffuse & $+0.026$ [$+0.007$, $+0.055$] & $+0.033$ [$+0.012$, $+0.064$] & 2 & +43\% \\
Floats or falls & 50-step, jump prompts & $-0.000$ [$-0.009$, $+0.008$] & $+0.001$ [$-0.009$, $+0.009$] & 2 & $-$0\% \\
Floats or falls & 1000-step, jump prompts & $+0.001$ [$-0.001$, $+0.003$] & $+0.003$ [$+0.000$, $+0.006$] & 1 & +3\% \\
Floats or falls & MotionDiffuse, jump prompts & $+0.001$ [$-0.017$, $+0.017$] & $+0.022$ [$-0.003$, $+0.054$] & 2 & +0\% \\
\addlinespace[2pt]
Fall rate (falling only) & 50-step & $+0.253$ [$+0.060$, $+0.505$] & $+0.408$ [$+0.178$, $+0.642$] & 2 & +0.122 \\
Fall rate (falling only) & 1000-step & $+0.082$ [$-0.035$, $+0.167$] & $+0.231$ [$+0.079$, $+0.409$] & 1 & +0.040 \\
Fall rate (falling only) & 50-step, jump prompts & $+0.077$ [$+0.051$, $+0.107$] & $+0.077$ [$+0.051$, $+0.107$] & 0 & +0.063 \\
Fall rate (falling only) & 1000-step, jump prompts & $+0.076$ [$+0.053$, $+0.096$] & $+0.082$ [$+0.060$, $+0.102$] & 1 & +0.056 \\
Fall rate (falling only) & MotionDiffuse, jump prompts & $+0.025$ [$+0.003$, $+0.047$] & $+0.034$ [$+0.009$, $+0.060$] & 2 & +0.020 \\
\bottomrule
\end{tabular}
}
\end{table}

\begin{table}[h]
\centering
\caption{Noise level at the recorded steps: $\bar\alpha_t$ and the signal-to-noise ratio $\bar\alpha_t / (1 - \bar\alpha_t)$ of the noisy input, from each model's schedule (MDM cosine, MotionDiffuse linear; \texttt{scripts/noise\_levels.py}).}
\label{tab:snr}
\begin{tabular}{lcccccccc}
\toprule
& \multicolumn{2}{c}{20\%} & \multicolumn{2}{c}{40\%} & \multicolumn{2}{c}{60\%} & \multicolumn{2}{c}{80\%} \\
\cmidrule(lr){2-3} \cmidrule(lr){4-5} \cmidrule(lr){6-7} \cmidrule(lr){8-9}
Model & $\bar\alpha$ & SNR & $\bar\alpha$ & SNR & $\bar\alpha$ & SNR & $\bar\alpha$ & SNR \\
\midrule
MDM, 50-step & 0.094 & 0.104 & 0.341 & 0.517 & 0.617 & 1.614 & 0.879 & 7.273 \\
MDM, 1000-step & 0.094 & 0.104 & 0.341 & 0.517 & 0.646 & 1.825 & 0.898 & 8.781 \\
MotionDiffuse & 0.002 & 0.002 & 0.026 & 0.027 & 0.194 & 0.240 & 0.656 & 1.910 \\
\bottomrule
\end{tabular}

\end{table}

\begin{table}[h]
\centering
\caption{Positive control (50-step model, 20\% of denoising, seeds pooled; gain with fallback over the tuned estimate, 95\% prompt-bootstrap intervals). Injected: the tuned estimate readout's out-of-fold residual plus noise, sized so that its best linear use would add the stated $R^2$ on the squared-error scale, added to the layer along one random direction with the variance of a typical direction of the standardized activations (\texttt{fair\_readouts.py -{}-layer-spec inject}). Tuned readouts can extract more than this linear amount. The real-layer rows come from the same rerun as the injected rows and differ from Table~\ref{tab:tuned} by refit noise of at most 0.001.}
\label{tab:control}
\resizebox{\linewidth}{!}{\begin{tabular}{llcc}
\toprule
& Added to the estimate & Pooled score & Within prompt and frame \\
\midrule
Dynamic balance ($R^2$) & the layer (real) & $+0.000$ [$-0.001$, $+0.001$] & $+0.000$ [$-0.001$, $+0.001$] \\
 & injected, 0.005 & $+0.013$ [$+0.009$, $+0.017$] & $+0.018$ [$+0.013$, $+0.022$] \\
 & injected, 0.01 & $+0.025$ [$+0.020$, $+0.030$] & $+0.035$ [$+0.031$, $+0.040$] \\
 & injected, 0.02 & $+0.040$ [$+0.034$, $+0.047$] & $+0.055$ [$+0.049$, $+0.061$] \\
\addlinespace[2pt]
In flight (AUC) & the layer (real) & $-0.004$ [$-0.008$, $+0.000$] & $-0.008$ [$-0.014$, $-0.002$] \\
 & injected, 0.005 & $-0.010$ [$-0.021$, $-0.002$] & $+0.018$ [$+0.009$, $+0.028$] \\
 & injected, 0.01 & $+0.001$ [$-0.006$, $+0.008$] & $+0.042$ [$+0.030$, $+0.056$] \\
 & injected, 0.02 & $+0.001$ [$-0.006$, $+0.008$] & $+0.072$ [$+0.057$, $+0.092$] \\
\addlinespace[2pt]
Floats or falls, jump prompts (AUC) & the layer (real) & $-0.000$ [$-0.008$, $+0.008$] & $-0.000$ [$-0.014$, $+0.016$] \\
 & injected, 0.005 & $+0.001$ [$-0.005$, $+0.007$] & $+0.006$ [$-0.014$, $+0.024$] \\
 & injected, 0.01 & $+0.002$ [$-0.008$, $+0.011$] & $+0.030$ [$+0.002$, $+0.059$] \\
 & injected, 0.02 & $+0.004$ [$-0.005$, $+0.012$] & $+0.057$ [$+0.030$, $+0.085$] \\
\bottomrule
\end{tabular}
}
\end{table}

\paragraph{Sample-level prediction.} Per-sample failure rates are predicted from frame-averaged activations with ridge regression ($\alpha = 10$; $\alpha = 100$ for the CLIP text embedding) in prompt-grouped 5-fold cross-validation. The layer used for the within-prompt comparison is chosen on the first seed by its mean within-prompt correlation and then evaluated on the other seeds.

\paragraph{Early-warning simulation.} For each prompt we pool its recorded samples across seeds and simulate 200 requests. A request draws a sample uniformly with replacement; an aborted attempt costs $k+1$ steps for a check at step $k$, a completed one costs the full number of steps; the last allowed attempt is always delivered. Compute at matched failure is interpolated along each strategy's curve over retry budgets 1 to 20, at the failure rates that finish-then-filter reaches with 2, 3 and 5 tries; larger budgets reach a floor set by prompts that float in nearly every sample.

\section{The fair comparison with a fixed MLP, per seed, and the pilot}
\label{app:fixedmlp}

\begin{figure}[h]
\centering
\includegraphics[width=\linewidth]{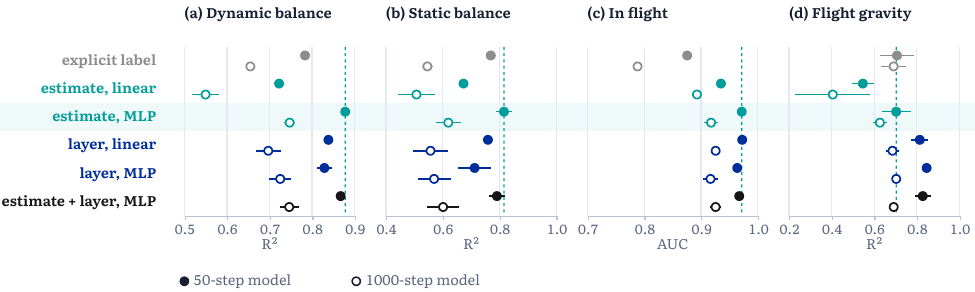}
\caption{Fair comparison at 20\% of denoising. Each row is a readout; filled circles show the 50-step model and open circles the 1000-step model (mean, with the standard deviation over seeds). The dashed line marks the MLP on everything the estimate shows (shaded row) for the 50-step model. For both balance measures it matches or beats every readout that uses the layer on both MDM checkpoints, and adding the layer to it (black) does not help; for flight this holds on the 50-step model, while on the 1000-step model the layer's linear probe is slightly ahead (0.926 against 0.918). Flight gravity is the exception in these data, on both checkpoints; Section~\ref{sec:fairresults} tests it with tuned readouts and about ten times as many falling frames.}
\label{fig:fair}
\end{figure}

\begin{table}[h]
\centering
\caption{Pooled scores at 20\% of denoising (mean $\pm$ standard deviation over seeds; best per row in bold). Estimate, MLP reads everything the estimate shows; Estimate + layer adds the layer's activations.}
\label{tab:fair}
\resizebox{\linewidth}{!}{%
\begin{tabular}{llcccccc}
\toprule
Model & Target & Explicit & Estimate, linear & Estimate, MLP & Layer, linear & Layer, MLP & Estimate + layer, MLP \\
\midrule
\multirow{4}{*}{50-step} & Dynamic balance ($R^2$) & 0.783 $\pm$ 0.009 & 0.723 $\pm$ 0.007 & \textbf{0.878 $\pm$ 0.005} & 0.838 $\pm$ 0.001 & 0.829 $\pm$ 0.018 & 0.867 $\pm$ 0.007 \\
 & Static balance ($R^2$) & 0.769 $\pm$ 0.019 & 0.673 $\pm$ 0.014 & \textbf{0.816 $\pm$ 0.028} & 0.759 $\pm$ 0.014 & 0.712 $\pm$ 0.059 & 0.791 $\pm$ 0.028 \\
 & In flight (AUC) & 0.876 $\pm$ 0.005 & 0.935 $\pm$ 0.008 & 0.972 $\pm$ 0.001 & \textbf{0.972 $\pm$ 0.003} & 0.964 $\pm$ 0.004 & 0.967 $\pm$ 0.008 \\
 & Flight gravity ($R^2$) & 0.707 $\pm$ 0.079 & 0.547 $\pm$ 0.053 & 0.703 $\pm$ 0.068 & 0.814 $\pm$ 0.040 & \textbf{0.846 $\pm$ 0.010} & 0.828 $\pm$ 0.038 \\
\midrule
\multirow{4}{*}{1000-step} & Dynamic balance ($R^2$) & 0.655 $\pm$ 0.008 & 0.550 $\pm$ 0.031 & \textbf{0.747 $\pm$ 0.014} & 0.697 $\pm$ 0.029 & 0.725 $\pm$ 0.026 & 0.746 $\pm$ 0.022 \\
 & Static balance ($R^2$) & 0.546 $\pm$ 0.019 & 0.508 $\pm$ 0.066 & \textbf{0.620 $\pm$ 0.043} & 0.557 $\pm$ 0.063 & 0.570 $\pm$ 0.058 & 0.601 $\pm$ 0.057 \\
 & In flight (AUC) & 0.788 $\pm$ 0.003 & 0.893 $\pm$ 0.001 & 0.918 $\pm$ 0.013 & \textbf{0.926 $\pm$ 0.002} & 0.917 $\pm$ 0.013 & 0.926 $\pm$ 0.010 \\
 & Flight gravity ($R^2$) & 0.691 $\pm$ 0.060 & 0.405 $\pm$ 0.176 & 0.627 $\pm$ 0.033 & 0.686 $\pm$ 0.030 & \textbf{0.703 $\pm$ 0.016} & 0.691 $\pm$ 0.013 \\
\bottomrule
\end{tabular}}

\end{table}

\begin{table}[h]
\centering
\caption{Within-prompt scores at 20\% of denoising: Pearson $r$ after subtracting each prompt's mean from target and prediction, so that only sample-specific information counts (mean $\pm$ standard deviation over seeds; best per row in bold).}
\label{tab:fairwithin}
\resizebox{\linewidth}{!}{%
\begin{tabular}{llcccccc}
\toprule
Model & Target & Explicit & Estimate, linear & Estimate, MLP & Layer, linear & Layer, MLP & Estimate + layer, MLP \\
\midrule
\multirow{4}{*}{50-step} & Dynamic balance ($r$) & 0.797 $\pm$ 0.011 & 0.756 $\pm$ 0.007 & \textbf{0.892 $\pm$ 0.003} & 0.851 $\pm$ 0.000 & 0.867 $\pm$ 0.002 & 0.889 $\pm$ 0.002 \\
 & Static balance ($r$) & 0.746 $\pm$ 0.020 & 0.644 $\pm$ 0.025 & \textbf{0.801 $\pm$ 0.023} & 0.729 $\pm$ 0.015 & 0.749 $\pm$ 0.020 & 0.795 $\pm$ 0.019 \\
 & In flight ($r$) & 0.710 $\pm$ 0.007 & 0.562 $\pm$ 0.014 & 0.755 $\pm$ 0.010 & 0.662 $\pm$ 0.010 & 0.734 $\pm$ 0.007 & \textbf{0.760 $\pm$ 0.008} \\
 & Flight gravity ($r$) & 0.439 $\pm$ 0.053 & 0.219 $\pm$ 0.047 & 0.340 $\pm$ 0.039 & 0.445 $\pm$ 0.071 & \textbf{0.536 $\pm$ 0.109} & 0.508 $\pm$ 0.081 \\
\midrule
\multirow{4}{*}{1000-step} & Dynamic balance ($r$) & 0.736 $\pm$ 0.004 & 0.664 $\pm$ 0.008 & 0.816 $\pm$ 0.001 & 0.772 $\pm$ 0.011 & 0.800 $\pm$ 0.006 & \textbf{0.817 $\pm$ 0.003} \\
 & Static balance ($r$) & 0.607 $\pm$ 0.010 & 0.572 $\pm$ 0.010 & \textbf{0.679 $\pm$ 0.003} & 0.610 $\pm$ 0.010 & 0.642 $\pm$ 0.010 & 0.673 $\pm$ 0.005 \\
 & In flight ($r$) & 0.562 $\pm$ 0.012 & 0.502 $\pm$ 0.020 & 0.608 $\pm$ 0.035 & 0.556 $\pm$ 0.024 & 0.580 $\pm$ 0.025 & \textbf{0.612 $\pm$ 0.034} \\
 & Flight gravity ($r$) & \textbf{0.645 $\pm$ 0.079} & 0.405 $\pm$ 0.101 & 0.576 $\pm$ 0.020 & 0.572 $\pm$ 0.023 & 0.614 $\pm$ 0.012 & 0.616 $\pm$ 0.003 \\
\bottomrule
\end{tabular}}

\end{table}

The first version of the fair comparison uses one fixed MLP (Section~\ref{sec:fair}) per seed. Figure~\ref{fig:fair} and Tables~\ref{tab:fair} and \ref{tab:fairwithin} replace the explicit label and the linear probe on the estimate with the same MLP that reads the layer. For both balance measures, the MLP on the estimate beats both the layer's linear probe and the same MLP on the layer, on both checkpoints; for dynamic balance on the 50-step model it scores 0.878, against 0.838 and 0.829. Adding the layer to the estimate changes $R^2$ by $-0.018$ to $-0.007$ for dynamic balance and by $-0.026$ to $-0.023$ for static balance on the 50-step model, and by $-0.012$ to $+0.005$ and $-0.037$ to $-0.002$ on the 1000-step model. For flight the readouts tie at an AUC of 0.97 on the 50-step model, and adding the layer changes the AUC by $-0.012$ to $+0.012$ across both MDM checkpoints. Within prompts (Table~\ref{tab:fairwithin}) the ordering is the same, so the conclusion does not rest on differences between prompts.

\begin{figure}[h]
\centering
\includegraphics[width=0.55\linewidth]{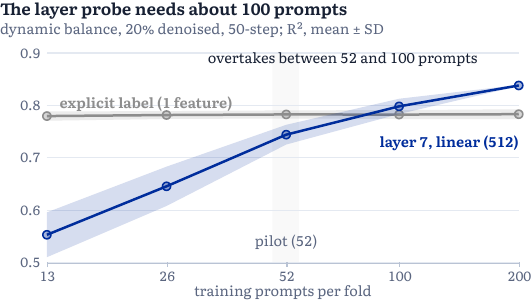}
\caption{Dynamic balance at 20\% of denoising (50-step model) with the probes trained on fewer prompts; same test folds; each point averages three draws of the training prompts in each of the three seeds (lines: mean; bands: standard deviation). The 512-feature layer probe overtakes the one-feature explicit label between 52 and 100 training prompts; the pilot (gray band) trained on 52.}
\label{fig:learning}
\end{figure}

Figure~\ref{fig:learning} shows why the pilot of Section~\ref{sec:learning} pointed the other way, with the linear layer probe of Section~\ref{sec:readable} trained on fewer prompts. The explicit label has a single feature and barely depends on training size (mean $R^2$ 0.780 to 0.783 across sizes). The linear layer probe has 512 features and improves steadily, from 0.553 with 13 training prompts and 0.744 with 52 to 0.798 with 100 and 0.837 to 0.839 with all of them. Applied to the larger data, the pilot's exact split gives gaps of $-0.003$, $+0.022$ and $+0.049$ across seeds, so its 12-prompt test set was also too small to fix the sign. Whether the layers appear to know more or less than the model shows can thus be set by the amount of training data, independently of what the model represents.

\section{The airborne label and the floating stretches}
\label{app:labelfix}

Our first airborne rule required only that every foot joint be above its contact height. On one 50-step run (1,000 motions) this gave 418 airborne stretches of at least 0.2\,s: 209 ballistic ($\hat g > 5\,\mathrm{m/s^2}$, 163 from the curated jump prompts) and 195 with $|\hat g| < 2\,\mathrm{m/s^2}$, 191 of them from other prompts. Inspecting the latter:

\begin{center}
\begin{tabular}{lr}
\toprule
Near-zero-gravity stretch & Stretches \\
\midrule
Supported by another body part (a wrist or knee within 15\,cm of the floor) & 48 \\
Feet raised but nearly still (for example a balance beam) & 48 \\
Whole motion above the floor & 5 \\
Unsupported and not falling & 94 \\
\bottomrule
\end{tabular}
\end{center}

The unsupported cases are ordinary walking poses with the whole body 7 to 12\,cm off the floor (median lowest foot 8.3\,cm, median duration 0.6\,s) in motions whose lowest foot elsewhere is at the floor (median 2.0\,cm): a real floating artifact. A foot-clearance threshold does not separate the groups: requiring the lowest foot to rise above 15\,cm keeps 73\% of ballistic stretches but also 44\% of near-zero ones, because cartwheels and crawling lift the feet. We therefore exclude stretches supported by a knee, wrist or the pelvis, and split the remaining unsupported stretches by whether they fall (Section~\ref{sec:labels}). Raised feet with no other support still count as airborne; a scene-free label cannot tell a balance beam from floating. All results in the main text use the fixed label. Only gravity-related labels changed; readouts that take the estimate's airborne flag as an input moved by at most 0.013.

\section{Sensitivity to the label thresholds}
\label{app:sensitivity}

\begin{table}[h]
\centering
\caption{Label thresholds changed one at a time (range over the three seeds of each model; the early rule ``abort if the decoded estimate shows floating'' is scored over the seeds of both MDM checkpoints).}
\label{tab:sensitivity}
\resizebox{\linewidth}{!}{\begin{tabular}{lcccccc}
\toprule
& \multicolumn{2}{c}{Floating, test captions (\% of frames)} & \multicolumn{2}{c}{Jump gravity, curated jump prompts (m/s$^2$)} & \multicolumn{2}{c}{Early rule at 20\%} \\
\cmidrule(lr){2-3} \cmidrule(lr){4-5} \cmidrule(lr){6-7}
Setting & 50-step & 1000-step & 50-step & 1000-step & Precision & Recall \\
\midrule
Default & 3.8--4.7 & 8.6--9.7 & 9.54--9.75 & 9.13--9.22 & 0.73--0.89 & 0.44--0.68 \\
Contact heights $\times 0.75$ & 6.4--7.6 & 13.1--13.8 & 9.68--9.92 & 9.03--9.40 & 0.66--0.78 & 0.46--0.58 \\
Contact heights $\times 1.25$ & 2.8--3.6 & 6.1--6.6 & 9.39--9.72 & 9.18--9.38 & 0.77--0.94 & 0.49--0.69 \\
Support rule 10\,cm & 3.8--4.7 & 8.7--9.7 & 9.54--9.75 & 9.13--9.22 & 0.72--0.91 & 0.45--0.65 \\
Support rule 20\,cm & 3.8--4.7 & 8.6--9.6 & 9.54--9.75 & 9.13--9.22 & 0.74--0.90 & 0.44--0.67 \\
Floating cut-off 1\,m/s$^2$ & 3.7--4.6 & 8.5--9.5 & 9.54--9.66 & 9.03--9.22 & 0.77--0.89 & 0.45--0.69 \\
Floating cut-off 3\,m/s$^2$ & 3.9--4.7 & 8.7--9.8 & 9.54--9.75 & 9.18--9.36 & 0.77--0.89 & 0.46--0.68 \\
\bottomrule
\end{tabular}
}
\end{table}

We recomputed every label from the cached joints with one threshold changed at a time: the ankle and toe contact heights scaled by 0.75 and 1.25, the support rule at 10 and 20\,cm instead of 15\,cm, and the floating cut-off at 1 and 3\,m/s$^2$ instead of 2\,m/s$^2$ (Table~\ref{tab:sensitivity}). Contact heights move the floating rates the most, since they decide which low poses count as unsupported; the support rule and the cut-off barely matter. Under every setting, every 1000-step seed floats more than every 50-step seed on test captions, by a factor of 1.9 to 2.1 in the means, and the median gravity of jumps on curated jump prompts stays between 9.03 and 9.92\,m/s$^2$. The training-free early rule keeps a precision of 0.66 to 0.94 and a recall of 0.44 to 0.69 across settings, seeds and models.

We also reran the fixed-MLP comparison of Appendix~\ref{app:fixedmlp} at 20\% of denoising under each setting, on the first seed of each model (dynamic balance only under the contact settings, since it depends on no other threshold). Its conclusions do not change. For dynamic balance, the MLP on the estimate beats the layer's linear probe by 0.020 to 0.059 in $R^2$, and adding the layer to it changes $R^2$ by $-0.018$ to $+0.004$, while the standard comparison favors the layer over the explicit label by 0.031 to 0.102. For flight gravity (layer 5), adding the layer gains 0.023 to 0.102, as with the default thresholds. The floating cut-off does not enter the probed labels, so those runs reproduce the default exactly.

\section{The labels on real motion capture}
\label{app:realmotion}

To check the labels where the physics is known, we ran them on the BMLmovi subset of AMASS \citep{mahmood2019amass,ghorbani2021movi}: 1,801 motion-capture clips of 86 people performing about 20 scripted everyday actions. Each clip goes through the steps HumanML3D uses to build its training data, so that real and generated motion are labelled in the same form: joint positions from the SMPL+H body model \citep{romero2017mano} with the clip's gender and the first 10 shape coefficients, downsampling to 20\,fps, HumanML3D's axis convention, retargeting to its uniform skeleton with MDM's own routine (the retargeted bone lengths match those of a HumanML3D processed clip exactly), and the floor at the lowest joint. The label code is unchanged.

Real people do not float, and the labels agree. Floating covers 0.09\% of frames and occurs in 4 of the 1,801 clips, against 3.8 to 4.7\% (50-step) and 8.6 to 9.7\% (1000-step) of frames on test captions for MDM; under the threshold settings of Appendix~\ref{app:sensitivity} it stays between 0.07 and 0.27\%. A raised foot in contact covers 0.9\% of contact frames, against 2.9 to 5.4\% for MDM. Of the 302 airborne stretches of at least 0.2\,s, 5 (1.7\%) are labelled floating.

The 297 falling stretches give the noise of the gravity label. At each subject's own scale their fitted gravity has a median of 9.65 and a standard deviation of 0.77\,m/s$^2$. HumanML3D's retargeting, however, scales each subject's root trajectory by the ratio of leg lengths (median 1.067, range 0.92 to 1.23), and fitted gravity scales with it: in the form MDM is trained on, real jumps fall at a median of 10.33\,m/s$^2$ (standard deviation 1.06). MDM's jumps, at 9.1 to 9.8\,m/s$^2$, therefore fall somewhat slower than those of its training data. Taking the frame-weighted spread of real jumps as label noise (0.68 to 0.98\,m/s$^2$, depending on whether the retargeting spread counts as noise), the fall rate of the jump-heavy runs, with a standard deviation of 2.24\,m/s$^2$, could be predicted up to $R^2 = 0.81$ to 0.91. The readouts of Section~\ref{sec:fairresults} reach 0.36 to 0.46 on the MDM checkpoints across refits, so the fall rate is hard to predict at 20\% of denoising for reasons other than label noise.

\section{Video test details}
\label{app:video}

The IntPhys development set \citep{riochet2018intphys} has 360 videos in three blocks (object permanence, shape constancy, spatio-temporal continuity), 30 scenarios per block, each with two possible and two impossible versions. We take the first 97 RGB frames of each video, resize them from $288 \times 288$ to $256 \times 256$ with antialiased bilinear interpolation, scale them to $[-1, 1]$ and encode them with the Wan~2.1 VAE \citep{wan2025}, keeping the posterior mean. On three videos, encoding on the laptop GPU matched the CPU to $1.4 \times 10^{-6}$ relative error, and decoding reconstructed the frames at 36 to 42\,dB PSNR.

Linear readouts are ridge classifiers (targets $\pm 1$) on standardized features, solved in the dual, with the penalty chosen from $\{10^{-3}, \dots, 10^{3}\}$ times the feature dimension by 4-fold cross-validation over the training scenarios. The change features $|\Delta|$ are the absolute differences between consecutive latent frames of the $4 \times 4$ grid features (for ``time kept + $|\Delta|$'', of the per-frame spatial means); the signed differences are the same without the absolute value, and therefore a linear function of the grid features. The temporal CNN applies two 1D convolutions (64 channels, kernel 3, ReLU, dropout 0.3) over the 25 latent frames of the grid features and their changes, followed by max- and mean-pooling over time and a linear output; it is trained with AdamW and early stopping on held-out training scenarios, averaged over three initializations. All splits keep a scenario's four videos together. The label-shuffle null permutes labels within each scenario, which keeps two possible and two impossible videos per scenario, and refits the linear readouts on the shuffled labels, 1,000 times for the three readouts with frame changes (20 times for the others).

\paragraph{Readouts tried.} We tried ten readouts of the clean latent, all reported in Table~\ref{tab:intphys} or here: the pooled baseline; time kept; time with the $4 \times 4$ grid; the same with signed changes; with absolute changes $|\Delta|$; $|\Delta|$ of the grid alone (77.9\% pairwise); time kept with the $|\Delta|$ of its spatial means (57.4\%); the temporal CNN; and two training-free scores, the largest global change and the largest change of a $4 \times 4$ cell (both 50.0\%). The absolute change is the only fixed nonlinearity we tried. The readouts with $|\Delta|$ of the grid are the best in each of the three blocks (78.5 to 82.0\% pairwise, against at most 64.8\% for any other readout), so choosing the readout on any one block and testing it on the other two gives the same result.

\section{Proofs}
\label{app:proofs}

\paragraph{Proposition~\ref{prop:vinfo}.} For a fixed regression function $f$, the expected log loss of $\mathcal N(f(X), \sigma^2)$ is $\tfrac12 \log(2\pi\sigma^2) + \mathbb E[(y - f(X))^2]/(2\sigma^2)$, minimized at $\sigma^2 = \mathbb E[(y-f(X))^2]$ with value $\tfrac12 \log\big(2\pi e\, \mathbb E[(y-f(X))^2]\big)$. Taking the infimum over the class, $H_{\mathcal V}(y \mid X) = \tfrac12 \log(2\pi e\, M^*_X)$ with $M^*_X = \inf_f \mathbb E[(y - f(X))^2] = (1 - R^2_X)\,\mathrm{Var}(y)$. The conditional usable information is the difference of the two entropies, in which $2\pi e\,\mathrm{Var}(y)$ cancels. Because the class on $(\hat\vx_0, \va)$ contains the class on $\hat\vx_0$, $M^*_{\hat\vx_0, \va} \le M^*_{\hat\vx_0}$, so the difference is nonnegative. $\square$

Two remarks on its use. The tuned readouts fit every target, including the AUC targets, by squared error; for those the identity holds for their mean squared error, not for the AUC we report. And a tuned combined readout can score below the estimate's on held-out prompts, because each is selected on a finite validation split. The readout with fallback chooses among the candidates of both input sets by the same validation error, so it restores the nesting of the proposition at the level of model selection; a negative measured gain can then come only from the difference between validation and test prompts, from selecting by squared error while scoring by AUC, and from refitting ridge after selection.

\paragraph{Proposition~\ref{prop:within}.} The within-cell score is the (frame-weighted) correlation of $\tilde y$ and $\tilde{\hat y}$, where $\tilde v = v - \bar v_c$ subtracts the mean of $v$ over the frames in the same cell. Centring is linear, and it maps any function of the cell to zero: $h(c)$ is constant within each cell and equals its own cell mean. Hence $\widetilde{f + h} = \tilde f$, and the two predictions have the same within-cell score. $\square$

The proposition covers additive effects only. A readout of $(X, u)$ may also use the text to change how it reads $X$, which can raise the within-cell score; the text control of Section~\ref{sec:fairresults} gives the readouts the text embedding itself, with families that can model such interactions.

\paragraph{Proposition~\ref{prop:ceiling}.} Fix a prompt $u$ and write $p = p_u$. The simulation draws samples of the prompt independently with replacement. Under finish-then-filter, attempt $j \in \{0, \dots, B-1\}$ is made if the $j$ earlier ones all failed, with probability $p^j$, and costs one full sample, so the expected cost is $\sum_{j<B} p^j$, and the delivered motion fails with probability $p^B$. A perfect detector with the final check kept makes the same attempts, since it aborts exactly the attempts that would fail and the last attempt is always finished. It therefore delivers the same failure $p^B$, and it pays $\tau$ instead of $1$ for each failed attempt before the last. Attempt $j < B-1$ is made and fails with probability $p^{j+1}$, so the expected saving is $(1-\tau)\sum_{j=1}^{B-1} p^j$. Summing numerator and denominator over prompts, each weighted equally as in the simulation, gives $S_B$.

For the second claim, fix any sequence of draws for a prompt. Let $j^*$ be the first passing draw. The perfect detector stops at $\min(j^*, B-1)$ and pays $\tau$ for each earlier draw and $1$ for the last. Any other detector that checks at the same step pays at least $\tau$ for every draw it makes and $1$ for the one it delivers. It cannot stop before $j^*$ unless it is at the last attempt, because every earlier draw fails, either when aborted or at the final check. So it makes at least as many draws, and pays at least as much. And the perfect detector delivers a failure only if all $B$ draws fail, in which case every detector does. Both inequalities hold for every sequence of draws, hence in expectation. $\square$

\end{document}